\documentclass[twoside,11pt]{article}

\usepackage[preprint]{jmlr2e}

\usepackage{amsmath,amsfonts,amssymb,booktabs}
\usepackage{enumerate,color,xcolor}
\usepackage{graphicx}
\usepackage{url}
\usepackage{todo} 
\usepackage{mathrsfs}
\usepackage{xfrac}
\usepackage{mathtools}
\usepackage{ulem}

\allowdisplaybreaks

\usepackage{caption}
\usepackage{subcaption}
\usepackage{graphicx}
\graphicspath{ {./images/} }

\usepackage{algorithm,algpseudocode}

\newtheorem{assumption}[theorem]{Assumption}
\usepackage{comment}

\def\<{\langle} \def\>{\rangle}
\def\eps{\varepsilon}

\def\div{\mathord{{\rm div}}}

\newcommand{\beq}{\begin{eqnarray}} 
	\newcommand{\eeq}{\end{eqnarray}}
\newcommand{\beqs}{\begin{eqnarray*}} 
	\newcommand{\eeqs}{\end{eqnarray*}}
\newcommand{\R}{\mathbb{R}}

\makeatletter
\def\munderbar#1{\underline{\sbox\tw@{$#1$}\dp\tw@\z@\box\tw@}}
\makeatother

\newcommand{\mg}[1]{{\color{magenta} #1}}
 
\usepackage{lastpage}  

\ShortHeadings{Anchored Langevin Algorithms}{G\"{u}rb\"{u}zbalaban, Nguyen, Zhang and Zhu} 
\firstpageno{1}

\renewcommand{\emph}[1]{\textit{#1}}

\begin{document}
	
\title{Anchored Langevin Algorithms}

\author{\name Mert G\"{u}rb\"{u}zbalaban \email mg1366@rutgers.edu \\
	\addr Department of Management Science and Information Systems\\
	Rutgers Business School\\
	Piscataway, NJ 08854, USA
	\AND
	\name Hoang M. Nguyen \email ngminhhoang7@gmail.com \\
	\addr Department of Mathematics\\
	Florida State University\\
	Tallahassee, FL 32306, USA
	\AND
	\name Xicheng Zhang \email XichengZhang@gmail.com \\
	\addr School of Mathematics and Statistics\\
	Beijing Institute of Technology\\
	Beijing 100081, P.R.China
	\AND
	\name Lingjiong Zhu \email zhu@math.fsu.edu \\
	\addr Department of Mathematics\\
	Florida State University\\
	Tallahassee, FL 32306, USA}

\editor{TBD}

\maketitle

\begin{abstract}
Standard first-order Langevin algorithms—such as the unadjusted Langevin algorithm (ULA)—are obtained by discretizing the Langevin diffusion and are widely used for sampling in machine learning because they scale to high dimensions and large datasets. However, they face two key limitations: (i) they require differentiable log-densities, excluding targets with non-differentiable components; and (ii) they generally fail to sample heavy-tailed targets. We propose \emph{anchored Langevin dynamics}, a unified approach that accommodates non-differentiable targets and certain classes of heavy-tailed distributions. The method replaces the original potential with a smooth reference potential and modifies the Langevin diffusion via multiplicative scaling. We establish non-asymptotic guarantees in the 2-Wasserstein distance to the target distribution and provide an equivalent formulation derived via a random time change of the Langevin diffusion. We provide numerical experiments to illustrate the theory and practical performance of our proposed approach.
\end{abstract}


\section{Introduction}

Sampling from a target probability distribution of the form \(\pi(x) \propto \exp(-U(x))\), where \(U: \mathbb{R}^d \to \mathbb{R}\), is a fundamental task with applications in statistics, machine learning, optimization, and operations research \citep{glasserman2004monte,gurbuzbalaban2022stochastic,bras2023langevin,lee2018convergence}. Markov Chain Monte Carlo (MCMC) methods---particularly those based on gradient-driven Langevin dynamics---have proven to be powerful tools for approximating samples from such distributions by exploiting the gradient \(\nabla U(x)\) to stochastically navigate the state space~\citep{RT96, teh2016consistency,welling2011bayesian}. 

Langevin-based MCMC algorithms are derived by discretizing diffusion processes that have $\pi$ as their stationary distribution. A notable example is the \emph{overdamped (or first-order) Langevin diffusion}:
\begin{equation}\label{overdamped:SDE}
dX_{t} = -\nabla U(X_{t})\,dt + \sqrt{2}\,dW_{t},
\end{equation}
where \(W_t\) denotes a standard \(d\)-dimensional Brownian motion initialized at zero. Indeed, under mild regularity conditions on \(U\), the stochastic differential equation (SDE) \eqref{overdamped:SDE} admits a unique stationary distribution with density \(\pi(x) \propto e^{-U(x)}\), commonly referred to as the \emph{Gibbs distribution}. Different discretizations of this SDE lead to different variants of Langevin algorithms. A prominent example is the Euler--Maruyama discretization, which leads to the unadjusted Langevin algorithm (ULA) defined by the iterations:
\begin{equation}\label{discrete:overdamped}
x_{k+1} = x_{k} - \eta \nabla U(x_{k}) + \sqrt{2\eta}\,\xi_{k+1},
\end{equation}
where \(\xi_k\) are independent and identically distributed (i.i.d.) standard Gaussian vectors in \(\mathbb{R}^d\), and \(\eta > 0\) is the stepsize. Beyond Euler--Maruyama, many other discretization schemes such as implicit and semi-implicit methods have also been explored, each yielding alternative Langevin-based, see e.g., \citep{li2019stochastic,hodgkinson2021implicit}.

Langevin algorithms, including ULA and its Metropolis-adjusted variants, have a rich history. While earlier analyses focused on asymptotic convergence to the target distribution, recent work has increasingly provided non-asymptotic performance guarantees, particularly under smoothness and growth conditions on \(U\); see \citep{Dalalyan,DM2017, DM2016,Durmus18,CB2018,DK2017,Barkhagen2021,Chau2019,Tao2021,BCESZ2022,Zhang2019,Chewi2024} and the references therein.
Despite these advances, when the potential \(U(x)\) is \emph{non-differentiable} or exhibits \emph{sublinear growth}, as in \emph{heavy-tailed} settings, classical Langevin methods face serious practical and theoretical limitations. 
In such settings, standard Langevin algorithms—based on discretizations of the overdamped Langevin diffusion—often become ineffective or fail to converge to the target. For example, the lack of differentiability in $U$ causes the gradient-based updates to be ill-defined, while heavy-tailed distributions challenge the exponential ergodicity and concentration behavior assumed in many convergence analyses and can lead to instability or divergence \citep{RT96,he2022heavy,he-transformed,he-mean-square-analysis}. These issues are not merely theoretical curiosities—they arise frequently in modern Bayesian inference problems, including Bayesian logistic regression with sparsity inducing priors \citep{vono2018sparse}, Bayesian learning problems with heavy-tailed priors \citep{agapiou-ht-priors}, robust Bayesian linear models with Student-\(t\) or Cauchy errors \citep{gagnon2023theoretical}, and Bayesian deep learning problems with non-smooth activation functions or heavy-tailed priors \citep{gurbuzbalaban2024penalized,castillo2024posterior,fortuin2022priors}. 

To extend the applicability of gradient-based Langevin algorithms to non-smooth densities and to certain heavy-tailed distributions within a unified framework, we propose a new class of methods which we call \emph{anchored Langevin algorithms}. This approach introduces a novel surrogate-guided sampling mechanism in which Langevin dynamics are anchored to a tractable reference potential \( U_0 \). The anchor enables stable sampling from a broad class of target distributions by allowing gradient-based updates even when the true potential \( U \) is non-differentiable or exhibits sublinear growth satisfying certain conditions. The core idea is to approximate \( U \) with a more regular surrogate \( U_0 \), chosen so that \( \nabla U_0(x) \) is well-defined and efficient to sample with ULA. Langevin dynamics is then simulated using \( \nabla U_0(x) \) in place of \( \nabla U(x) \), while the discrepancy between \( U \) and \( U_0 \) is handled through a correction mechanism that appropriately scales both the injected Gaussian noise and the surrogate gradient. This results in a more general sampling framework than ULA and an efficient algorithm that retains the advantages of gradient-based sampling while overcoming some of the computational challenges posed by the original target distribution. Our contributions are as follows:

    First, we introduce the \emph{anchored Langevin SDE} (AL-SDE), which modifies both the drift and diffusion terms of the overdamped Langevin SDE using a reference (anchoring) potential \(U_0(x)\), leading to a state-dependent diffusion term. In Theorem~\ref{thm:continuous:time}, we show that, under suitable conditions on the modified drift, the AL-SDE admits \(\pi\) as its unique stationary distribution.
We then present several examples illustrating that, with appropriate choices of \( U_0(x) \), the AL-SDE can effectively sample from both light-tailed and heavy-tailed distributions, including the student-\( t \) distribution. Furthermore, we show that if \( \pi \) satisfies a Poincaré inequality, then the distribution \( \mu_t \) of the AL-SDE at time \( t \) converges exponentially fast in time $t$ to \( \pi \), provided that \( U_0(x) \) uniformly approximates \( U(x) \); that is, if \( \sup_{x\in\mathbb{R}^{d}} |U(x) - U_0(x)| \) is finite.

   In our second set of contributions, we construct strong solutions to the AL-SDE using a time-change argument. Specifically, we show that the AL-SDE can be interpreted as an overdamped Langevin SDE with potential \(U_0\) evaluated at a particular random time \(\ell(t)\), for which we provide an explicit expression. We then analyze Euler--Maruyama discretizations of the AL-SDE and, under suitable technical conditions on its drift and noise coefficients, establish non-asymptotic performance bounds on the 2-Wasserstein distance between the distribution of the iterates \(x_k\) in \eqref{discrete:dynamics} and the target distribution \(\pi\). A key challenge arises from the state-dependent diffusion coefficient \( \sigma(x_k) \) in the AL-SDE, which prevents the use of standard synchronous coupling arguments. Instead, we leverage the mean-square analysis framework developed in \cite{Tao2021}, which is well suited for systems with state-dependent diffusion terms.
    
    Third, we consider non-smooth potentials \( U(x) \) and construct smoothed approximations \( U_0(x) \) by convolving \( U \) with a mollifier—specifically, an infinitely divisible density \( \rho_\varepsilon \) that approximates \( U \) increasingly well as \( \varepsilon \to 0 \). We provide examples in which our drift conditions are satisfied under such smoothing. We then focus on the special case of Gaussian smoothing, where scaled Gaussian densities are used as mollifiers. For composite objectives of the form \( U(x) = f(x) + g(x) \), where \( f \) is smooth and strongly convex, and \( g \) is a non-smooth (possibly non-convex) but Lipschitz-continuous penalty function, we show that the proposed anchored Langevin algorithms can sample efficiently from the target distribution. These results demonstrate that, within our framework, the anchoring potential can be chosen as a smoothed version of the original non-smooth potential under some technical conditions.
    
Fourth, we demonstrate the performance of our method across a diverse set of problems. First, we simulate both univariate and multivariate Laplace distributions, which are characterized by non-smooth densities. Next, we consider sparse Bayesian logistic regression problems involving non-smooth priors such as SCAD, MCP, and mixed \( \ell_2 \)-\( \ell_1 \) penalties. In addition, we test the algorithm on a two-layer neural network, where the first layer uses a ReLU activation and the second layer uses a sigmoid function. Finally, we evaluate our method on a heavy-tailed target distribution with polynomial decay, where we show that anchored Langevin algorithms outperform the standard overdamped Langevin algorithm in sampling from such heavy-tailed distributions.
\section{Related Work}

For non-differentiable target distributions that are not heavy-tailed, zeroth-order Langevin algorithms can be employed. These methods approximate first-order information based on evaluations of the potential function $U$ using finite-differences \citep{roy2022stochastic,dwivedi2019log}. This can be beneficial in some settings when the gradients do not exist or when they are hard to compute. However, zeroth-order methods are typically slower than first-order methods, as their gradient estimates tend to be noisier. 

There are also alternative approaches that rely on approximating the potential with a smooth, differentiable surrogate to enable gradient-based sampling. Among these, ~\cite{Zhou14} propose a gradient-based adaptive stochastic search method that smooths the original objective by integrating it against a parameterized family of exponential densities, producing a differentiable surrogate. Additionally, proximal MCMC methods approximate the non-smooth function \( U \) using its Moreau--Yoshida envelope (MYE)~\citep{Durmus18,Goldman21,Mou19,Salim20,Lamperski20,Bernton18,Wibisono19,Pereyra2016,Eftekhari2023}, which provides a smooth surrogate. The MYE of a function \( U : \mathbb{R}^d \to \mathbb{R} \), defined as
$
U^{\lambda}(x) := \inf_{z\in\mathbb{R}^{d}} \left[ U(z) + \frac{1}{2\lambda} \|x - z\|^2 \right],
$
yields a smooth approximation \( U^\lambda \) that can be used with gradient-based Langevin algorithms for sampling. However, computing the gradient of the MYE requires evaluating a proximal step with respect to \(U\), which is typically computationally expensive \citep{Goldman21}, with efficient computations available only in specific cases.
Moreover, using the gradient of \( U^{\lambda} \) introduces bias and in some cases, a very tight envelope (i.e., a small \( \lambda \)) may be needed to obtain an accurate approximation. This, however, necessitates small stepsizes, leading to slow mixing \cite[Example 4.1]{Goldman21}. 

Other envelopes that approximate a non-smooth potential with a smooth one, such as the forward-backward envelope, can also be used \citep{Eftekhari2023}; but they still require computing proximal steps. Piecewise-deterministic Markov processes (PDMPs) such as the bouncy particle and zig-zag samplers, which do not suffer from asymptotic bias, offer an alternative. They can be applied to target distributions that are differentiable almost everywhere with potentially heavy tails \citep{deligiannidis2019exponential,durmus2020geometric,bierkens2019ergodicity}, and in practice, may be preferable to MYE-based methods—particularly when the MYE is difficult to compute or not well-defined 
\citep{Goldman21}. However, PDMPs may encounter computational difficulties in high dimensions; event-time simulations can be demanding, and their performance can be sensitive to the availability of tight event-rate bounds \citep{Goldman21}. Furthermore, to our knowledge, PDMPs do not admit non-asymptotic performance guarantees in the context of heavy-tailed sampling; existing guarantees have an asymptotic nature. 

An alternative strategy for handling non-smoothness is Gaussian smoothing \citep{nesterov2017random} where one would obtain a smooth approximation of $U$ by convolving it with a Gaussian kernel. \cite{Chatterji20} analyze Gaussian-smoothing-based Langevin dynamics for convex and non-smooth $U$. This approach replaces \(\nabla U\) in the Langevin SDE \eqref{overdamped:SDE} with \(\nabla U_0\) and then simulates the dynamics, which, as the discretization parameter vanishes, converges to an approximate target density \(\pi_0(x) \propto e^{-U_0(x)}\) and thus suffers from bias. By contrast, in our framework, since we modify the Langevin SDE itself, the resulting dynamics converge to the original target \(\pi(x) \propto e^{-U(x)}\) and are free of such bias. Moreover, our approach does not require assuming light-tailed target distributions. There also exist mirror-descent–type algorithms that employ the Bregman–Moreau envelope instead of the MYE to handle non-smoothness on Riemannian manifolds \citep{Lau2022}; but they suffer from similar computational drawbacks.

When $U$ is non-differentiable but convex, there are also subgradient-based approaches, which relaxes the differentiability requirement. Among subgradient-based approaches, \cite{Durmus19} proposed the Stochastic Subgradient Langevin Dynamics (SSGLD) and provided convergence guarantees by leveraging the fact that sampling can be viewed as optimization in the space of probability measures. Other subgradient-based approaches include \citep{habring2024subgradient}. However, to the best of our knowledge, these approaches do not extend to general heavy-tailed distributions, and their theory is restricted to convex potentials in both continuous- and discrete-time settings. By contrast, our framework guarantees convergence to the target distribution in continuous time without requiring convexity.

Sampling from target distributions with heavy tails—where the negative log-density \( U(x) \) grows sublinearly—presents unique challenges for standard sampling algorithms and the literature is quite limited. Classical Langevin algorithms, including the Unadjusted Langevin Algorithm (ULA) and Metro\-polis-Adjusted Langevin Algorithm (MALA), typically assume that \( U(x) \) grows at least linearly or faster to ensure geometric ergodicity and proper control over tail behavior. When \( U(x) = o(\|x\|) \) as $\|x\|\rightarrow\infty$, the resulting target distribution decays more slowly than exponentially, and standard Langevin algorithms may exhibit poor convergence or fail to explore the tails altogether \citep{RT96}. To address this, several algorithmic modifications have been proposed including works by \cite{Kamatani2017},
\cite{belomestny2021fourier}, \cite{bell2024adaptive}; however, these approaches rely on Metropolis steps, which can be expensive for some applications involving high dimensionality and large datasets \cite{welling2011bayesian}.
An interesting work by \cite{he-mean-square-analysis} develops zeroth- and first-order Langevin algorithms for heavy-tailed distributions whose potentials satisfy weighted Poincar\'e inequalities, including \(t\)-distributions. Their first-order method discretizes associated It\^o diffusions and extends to zeroth-order variants via Gaussian smoothing. However, when a non-smooth potential \(U\) is approximated by a smoothed \(U_0\), the limiting It\^o diffusions converge to \(\pi_0(x) \sim e^{-U_0(x)}\), leading to asymptotic bias. The modified diffusion we propose can sample \(\pi(x) \sim e^{-U(x)}\) directly without introducing a bias. In related work, \cite{he-transformed} characterize the class of heavy-tailed densities for which polynomial-order complexity guarantees can be obtained when the Unadjusted Langevin Algorithm is applied to suitably transformed versions of the target. They provide a precise characterization of smooth heavy-tailed densities that admit polynomial oracle complexity bounds in both dimension and inverse accuracy. Our framework is complementary to \cite{he-transformed}: it generalizes ULA dynamics to handle non-smooth targets, thereby extending the range of distributions from which efficient sampling is possible, while also offering a unified approach for heavy-tailed sampling. 


\subsection{Overdamped Langevin SDE}


The first non-asymptotic result of the discretized Langevin diffusion \eqref{discrete:overdamped}
is due to \cite{Dalalyan}, which was improved soon after
by \cite{DM2017} with a particular emphasis on the dependence on the dimension $d$. 
Both works consider the total variation as the distance to measure the convergence. 
Later, \cite{DM2016} studied the convergence in the 2-Wasserstein distance,
and \cite{Durmus18} studied variants of \eqref{discrete:overdamped} when $U$
is not smooth. 
\cite{CB2018} studied the convergence in the Kullback-Leibler (KL) distance.
\cite{DK2017} studied the convergence when only stochastic gradients are available. More recent studies include \cite{Barkhagen2021,Chau2019,Tao2021,BCESZ2022,Zhang2019,Chewi2024}.

In \eqref{overdamped:SDE}, we assume that $U:\mathbb{R}^d\to\mathbb{R}$ is a $C^1$-potential function with
$M:=\int_{\mathbb{R}^d} e^{-U(x)}dx<\infty$.
Since $x\mapsto\nabla U(x)$ is continuous, it is well known that for each starting point $x\in\mathbb{R}^d$,
the SDE in \eqref{overdamped:SDE} admits a unique weak solution $X_t(x)$ up to the explosion time $\zeta$ (see e.g. \cite{SV79}). If we further assume that
for some $c_0\in\mathbb{R}$ and $c_1>0$,
\begin{align}\label{Dis}
	-\<x,\nabla U(x)\>\leq c_0\Vert x\Vert^2+c_1,\ \ x\in\mathbb{R}^d,
\end{align}
then there is no explosion, i.e., $\zeta=\infty$ a.s. The semigroup associated with $X_t(x)$ is defined by
$P_tf(x):=\mathbb{E} f(X_t(x))$, $t>0$.
It is easy to check that the probability measure
$\pi(dx)= e^{-U(x)}dx/M$
is an invariant probability measure of $P_t$, 
i.e., for any $f\in C_b(\mathbb{R}^d)$,
$\int_{\mathbb{R}^d} P_tf(x)\pi(dx)=\int_{\mathbb{R}^d}f(x)\pi(dx)$.
Furthermore, if $c_0<0$ in \eqref{Dis}, then $\pi$ is the unique invariant probability measure and for some $C,\lambda>0$,
\begin{align}\label{ERG}
	|P_tf(x)-\pi(f)|\leq C e^{-\lambda t}.
\end{align}
The classical Markov Chain Monte Carlo (MCMC) method is based on using the distribution of $X_t$ to approximate $\pi$ when $t\to\infty$.

However, it is well known that the exponential convergence \eqref{ERG} does not hold for $U(x)=\Vert x\Vert^{\gamma}$ with $\gamma\in(0,1)$ 
(see \cite[Theorems 2.3, 2.4]{RT96}).
Therefore, it is not expected that one can use SDE \eqref{overdamped:SDE} to simulate the heavy-tailed distribution $\mu$ 
such as $U(x)=\beta\log(1+\Vert x\Vert^2)$ with $\beta>\frac{d}{2}$, that is, the invariant measure $\pi \propto (1+\Vert x\Vert^2)^{-\beta}$.

\subsection*{Notations}

We summarize the notations here for readers' convenience.
\begin{itemize}
\item For a differentiable function $f:\mathbb{R}^d\to\mathbb{R}$, $\nabla f:=(\partial_1 f,\cdots,\partial_d f)$.
\item For two vectors $a,b\in\mathbb{R}^d$, we use $\<a,b\>$ to denote the inner product.
\item For a vector $x\in\mathbb{R}^{d}$, let $\Vert x\Vert:=\sqrt{\langle x,x\rangle}$ be the $\ell_{2}$-norm.
For a matrix $A\in\mathbb{R}^{d\times d}$, let $\Vert A\Vert_{\mathrm{HS}}:=\sqrt{\text{Tr}(AA^{\top})}$ be the Hilbert-Schmidt norm.
    \item For a bounded measurable function $f:\mathbb{R}^d\to\mathbb{R}$, $\|f\|_\infty:=\sup_{x\in\mathbb{R}^d}|f(x)|$.
    \item For a signed measure $\mu$ on $\mathbb{R}^d$, 
    $\|\mu\|_{\rm var}:=\sup_{\|\varphi\|_\infty\leq 1}\mu(\varphi).$
\item $\mathbb{W}$ denotes the space of all continuous functions from $[0,\infty)$ to $\mathbb{R}^d$, which is endowed with the topology of locally uniform convergence.
\item
For any two real numbers $x,y$, we denote
$x\vee y:=\max\{x,y\}$
and $x\wedge y:=\min\{x,y\}$.
\end{itemize}

\section{Anchored Langevin SDE}

Let $U, U_0:\mathbb{R}^d\to\mathbb{R}$ be two continuous functions.
Consider the SDE:
\begin{equation}\label{SDE0}
dX_{t}=b(X_{t})dt+\sqrt{2}\sigma(X_{t})dW_{t},   
\end{equation}
where we define the drift term and the diffusion term as
\begin{align}\label{11}
	b(x):=-\nabla U_0(x) e^{(U-U_0)(x)},\ \ \sigma(x):= e^{(U-U_0)(x)/2},
\end{align}
where $U_{0}:\mathbb{R}^{d}\rightarrow\mathbb{R}$
plays the role as a reference potential.
Therefore, the dynamics of the overdamped
Langevin SDE \eqref{overdamped:SDE} 
is modified so that
it is \emph{anchored} with a new potential $U_{0}$:
\begin{equation}\label{overdamped:SDE:modified}
	dX_{t}=-\nabla U_{0}(X_{t})e^{U(X_{t})-U_{0}(X_{t})}dt
	+\sqrt{2}e^{(U(X_{t})-U_{0}(X_{t}))/2}dW_{t},
\end{equation}
and we name the SDE \eqref{overdamped:SDE:modified}
the \emph{anchored Langevin SDE}. 

It is well known that the distribution of the overdamped Langevin SDE $X_{t}$
given in \eqref{overdamped:SDE} converges to a unique invariant distribution,
with density $\pi(x)\propto e^{-U(x)}$, 
which is known as the Gibbs distribution.
One can show that the modified SDE \eqref{overdamped:SDE:modified}
with the anchored reference potential $U_{0}$
preserves the Gibbs distribution $\pi\propto e^{-U(x)}$
as an invariant distribution.

\begin{assumption}\label{assump:continuous}
    Suppose that for some $c_0,c_1>0$ and $r>-1$,
	\begin{align}\label{CN1}
		[d-\<x, \nabla U_0(x)\>] e^{(U-U_0)(x)}\leq -c_0\Vert x\Vert^{2+r}+c_1.
	\end{align}
\end{assumption}

Under \eqref{CN1}
and the assumptions $U_0\in C^2$ and $U\in C^1$, 
there is a unique strong solution to SDE \eqref{SDE0} (see e.g. \cite{GK96,G98}).
Let $P_t$ be the semigroup defined by the anchored Langevin SDE \eqref{SDE0}.
We have the following result.

\begin{theorem}\label{thm:continuous:time}
	Under Assumption~\ref{assump:continuous}, $\pi$ is the unique invariant measure of the semigroup $P_t$. Moreover, 
	\begin{enumerate}[(i)]
		\item If $r\geq 0$, then there are $\lambda, C>0$ such that for all $t>0$ and $x\in\mathbb{R}^d$,
		$$
		\sup_{\|\varphi/V\|_\infty\leq 1}|P_t\varphi(x)-\pi(\varphi)|\leq C e^{-\lambda t}V(x),
		$$
		where $V(x):=1+\Vert x\Vert^2$.
		\item If $r>0$, then there are $\lambda, C>0$ such that for all $t>0$,
		$$
		\sup_{x\in\mathbb{R}^d}\|P_t(x,\cdot)-\pi\|_{\rm Var}\leq C e^{-\lambda t}.
		$$
	\end{enumerate}
\end{theorem}

It is known that the classical overdamped Langevin SDE \eqref{overdamped:SDE} fails to sample
heavy-tailed distributions with exponential ergodicity, i.e. it does not converge to the target exponentially fast in time; hence, even if it does converge, the convergence can be slow; see \cite{RT96}. 
In the following (Theorem~\ref{thm1}), we will show that the anchored Langevin SDE \eqref{SDE0}
can sample a \textit{heavy-tailed} Gibbs distribution $\pi\propto e^{-U(x)}$ with
convergence being exponentially fast in time, 
which covers the multivariate Student-$t$ distribution (Example~\ref{ex:student}).


\begin{theorem}\rm(Heavy-tailed distribution)\label{thm1}
Let $q:\R^d\to[1,\infty)$ be a $C^1$-function such that for some $\beta>0$, $$\int_{\R^d}q(x)^{-1-\beta}dx<\infty;$$
and for some $c_0,c_1>0$ and $r>-1$, and for all $x\in\R^d$,
$$
dq(x)-\beta\<x,\nabla q(x)\>\leq -c_0\|x\|^{2+r}+c_1.
$$
Let us choose 
\begin{equation*}
U_0(x):=\beta\log q(x),
\quad\text{and}\quad 
U(x):=(\beta+1)\log q(x).
\end{equation*}
Then Assumption~\ref{assump:continuous} is satisfied and
the anchored Langevin dynamics \eqref{overdamped:SDE:modified} has the unique stationary distribution 
$q(x)^{-1-\beta}/\int_{\R^d}q(x)^{-1-\beta}dx$.
\end{theorem}

Next, we will show that Theorem~\ref{thm1} covers
examples such as the multivariate Student-$t$ distribution.

\begin{example}\label{ex:student}
Consider the multivariate Student-$t$ distribution on $\mathbb{R}^d$ with $\nu> 1$ degrees of freedom, location parameter $\mu\in\mathbb{R}^d$, and symmetric positive definite scale matrix $\Sigma\in\mathbb{R}^{d\times d}$, with density
\[
\pi(x)\ \propto\ \Bigl(1+\tfrac{1}{\nu}(x-\mu)^\top\Sigma^{-1}(x-\mu)\Bigr)^{-(\nu+d)/2}.
\]
For $\alpha>1/2$,
let $q(x):=(1+\tfrac{1}{\nu}(x-\mu)^\top\Sigma^{-1}(x-\mu))^\alpha$. 
For $\beta:=\frac{d+\nu}{2\alpha}-1>\frac{d}{2\alpha}$, we have
\begin{align*}
dq(x)-\beta\<x,\nabla q(x)\>
&=\left[d-\frac{2\alpha\beta\<x,\Sigma^{-1}(x-\mu)\>}{\nu+(x-\mu)^\top\Sigma^{-1}(x-\mu)}\right]q(x)\\
&=\left[d-2\alpha\beta+\frac{2\alpha\beta(\nu-\<\mu,\Sigma^{-1}(x-\mu)\>)}{\nu+(x-\mu)^\top\Sigma^{-1}(x-\mu)}\right]q(x).
\end{align*}
Let $\lambda_{\rm min}$ be the minimum eigenvalue of $\Sigma^{-1}$. Noting that
$$
q(x)\geq \left(1+\lambda_{\rm min}\|x-\mu\|^2/\nu\right)^\alpha,
$$
it is easy to see that for some $0<c_0<(2\beta\alpha-d)(\lambda_{\rm min}/\nu)^\alpha$ and $c_1>0$,
$$
dq(x)-\beta\<x,\nabla q(x)\>\leq -c_0\|x\|^{2\alpha}+c_1.
$$
Consequently, by Theorem~\ref{thm1}, Assumption~\ref{assump:continuous} is satisfied with some $c_0,c_1>0$ and a parameter $r=2\alpha-2>-1$; in particular, the anchored Langevin SDE admits $\pi$ as its unique invariant measure and 
if $\alpha\geq 1$ (hence $\nu>2$), converges exponentially fast ($\alpha\geq 1$ is due to $r\geq 0$ in Theorem~\ref{thm:continuous:time} and $\nu>2$ is due to $\beta=\frac{d+\nu}{2\alpha}-1>\frac{d}{2\alpha}$ so that $\nu>2\alpha\geq 2$.). We note that multivariate Student-$t$-distributions have a finite mean only for $\nu>1$ and finite variance only for $\nu>2$, and many practical applications involve the $\nu>2$ case \citep{gelman2013bayesian}.
\end{example}

\begin{remark}\label{ex:light:tail}\rm(Light-tailed distribution)
    In Theorem~\ref{thm1} and Example~\ref{ex:student}, we showed that anchored Langevin SDE \eqref{SDE0}
can sample a heavy-tailed Gibbs distribution. Indeed, anchored Langevin SDE \eqref{SDE0}
can also be used to sample light-tailed Gibbs distributions.
	Consider the following example. 
    Let $\beta>0$ and
	$U_0(x):=(1+\Vert x\Vert^2)^{\beta/2}$.
	Suppose that for some $r_1\geq r_0\geq 1-\tfrac\beta 2$, $K\geq 0$ and all $\Vert x\Vert\geq K$, $U$ satisfies
	$r_0\log(1+\Vert x\Vert^2)\leq (U-U_0)(x)\leq r_1\log(1+\Vert x\Vert^2)$,
such that for all $\Vert x\Vert\geq K$,
	$$
	\left(1+\|x\|^2\right)^{-r_1} e^{-(1+\|x\|^2)^{\beta/2}}\leq  e^{-U(x)}\leq \left(1+\|x\|^2\right)^{-r_0} e^{-(1+\|x\|^2)^{\beta/2}}.
	$$
	In other words, $\pi\propto e^{-U}$ has light tails.
	As above one can check that for some $K'\geq K$, condition \eqref{CN1} in Assumption~\ref{assump:continuous} holds for all $\Vert x\Vert\geq K'$, and in this case,
	$$
    b(x)=-\beta x(1+\Vert x\Vert^2)^{\frac\beta 2-1} e^{(U-U_0)(x)},\ \ \sigma(x)= e^{(U-U_0)(x)/2},
	$$
	and for $\Vert x\Vert\geq K$, 
	$\|b(x)\|\leq 2\beta(1+\Vert x\Vert^2)^{r_1+\frac{\beta-1}2}$ and $|\sigma(x)|\leq (1+\Vert x\Vert^2)^{\frac{r_1}{2}}$.
    By Theorem~\ref{thm:continuous:time}, the anchored Langevin SDE \eqref{SDE0}
can sample $\pi\propto e^{-U(x)}$.
\end{remark} 


There are many desirable properties
of the anchored Langevin SDE.
As we have seen in previous discussions, 
one advantage of the anchored Langevin SDE \eqref{overdamped:SDE:modified}
is that it can target the Gibbs distribution $\pi$
when $\pi$ has heavy tails even though the overdamped Langevin SDE \eqref{overdamped:SDE}
cannot. 
In Theorem~\ref{thm:continuous:time}, we showed that 
the anchored Langevin SDE \eqref{overdamped:SDE:modified} converges
exponentially fast in $t$ to the same Gibbs distribution $\pi$.
Next, we obtain a complementary result to Theorem~\ref{thm:continuous:time},
that shows the convergence in $\chi^{2}$-divergence.
Let $\mu_{t}$ denote the distribution of the anchored Langevin SDE $(X_{t})_{t\geq 0}$ 
in \eqref{overdamped:SDE:modified}. To quantify the 
convergence of $\mu_{t}$ to the Gibbs distribution $\pi$, 
we consider the $\chi^{2}$-divergence:
\begin{equation}
	\chi^{2}(\mu_{t}\Vert\pi)=\int_{\mathbb{R}^{d}}\left(\frac{d\mu_{t}}{d\pi}-1\right)^{2}d\pi,
\end{equation}
and before we proceed, let us introduce
the following technical lemmas.

\begin{lemma}\label{prop:reversible}
	Under Assumption~\ref{assump:continuous},
    the anchored Langevin SDE \eqref{overdamped:SDE:modified} is reversible. 
\end{lemma}

\begin{lemma}\label{lem:Dirichlet}
Let $\mathcal{E}(f):=-\int_{\mathbb{R}^{d}}f\mathcal{L}(f)d\pi$
be the Dirichlet form. Then, we have
	\begin{equation}
		\mathcal{E}(f)=\int_{\mathbb{R}^{d}}e^{U-U_{0}}\left\Vert\nabla f\right\Vert^{2}d\pi.
	\end{equation}
\end{lemma}

Now, we are ready to state the following result, that characterizes
the convergence of the anchored Langevin SDE \eqref{overdamped:SDE:modified}
to the Gibbs distribution in $\chi^{2}$-divergence.

\begin{proposition}\label{prop:chi:square}
Suppose that Assumption~\ref{assump:continuous} holds.
If $\pi$ satisfies a Poincar\'{e} inequality with constant $C_{P}$, then
\begin{equation}
	\chi^{2}(\mu_{t}\Vert\pi)\leq
	\chi^{2}(\mu_{0}\Vert\pi)e^{-2at/C_{P}},
\end{equation}
provided that
$a:=e^{\inf_{x\in\mathbb{R}^{d}}(U(x)-U_{0}(x))}\in(0,\infty)$.
\end{proposition}

\begin{remark}
Note that Poincar\'{e} inequality may not hold for polynomial tails, in which case Proposition \ref{prop:chi:square} would not be applicable. However, Theorem~\ref{thm:continuous:time} relies instead on the problem-tailored Lyapunov drift and may still apply to polynomial tails.
\end{remark}



\subsection{Random time change}\label{time:change:section}
In this subsection we use the random time change to construct a solution of SDE \eqref{SDE0}. Let $Z_t$ solve the following overdamped Langevin SDE:
\begin{align}\label{SDE1}
	Z_t=X_0-\int^t_0\nabla U_0(Z_s)ds+\sqrt{2}\widetilde W_t,
\end{align}
where $\widetilde W_t$ is another $d$-dimensional standard Brownian motion.  It is well known that there is a unique strong solution to the above SDE. More precisely, 
there is a continuous functional $\Phi: \mathbb{W}\to \mathbb{W}$   so that
$Z_t=\Phi(\widetilde W)(t)$.

\begin{assumption}\label{assump:random:time}
Suppose that for some $c_0, K\geq 0$,
\begin{align}\label{DIS}
d-\<x, \nabla U_0(x)\>\leq-c_0,\quad \Vert x\Vert^2\geq K.
\end{align}    
\end{assumption}

For $t>0$, define
$$
\ell(t)=\left\{s>0: \int^s_0 e^{(U_0-U)(Z_r)}dr>t\right\}.
$$
We have the following technical lemma. 

\begin{lemma}\label{lem:crucial}
Under \eqref{DIS}, $\ell(t):[0,\infty)\to[0,\infty)$ is continuous differentiable  and solves
the following ordinary differential equation (ODE):
\begin{equation}\label{ODE1}
\frac{d\ell(t)}{dt}= e^{(U-U_{0})(Z_{\ell(t)})},\ \ell(0)=0.
\end{equation}
\end{lemma}

Now we are ready to state the following theorem, 
which states that anchored Langevin SDE $X_{t}$ can be viewed 
as an overdamped Langevin SDE $Z_{t}$ with target $U_0$ at random time $\ell(t)$. 

\begin{theorem}\label{thm:time:change}
Under \eqref{DIS}, 
$X_t:=Z_{\ell(t)}$ is the unique weak (strong) solution of SDE \eqref{SDE0}.
\end{theorem}

The main problem is how to simulate the solution of ODE \eqref{ODE1}.
For example, if $U_0(x)=\Vert x\Vert^2$, then $Z_t$ is an Ornstein-Uhlenbeck process. Thus, one can only simulate the solution of ODE \eqref{ODE1}.
This could be done via Euler-Maruyama discretizations, 
which will be introduced and studied in detail
in the next section.

\section{Anchored Langevin Dynamics}

\subsection{Non-asymptotic analysis for anchored Langevin dynamics}\label{sec:analysis}

We consider the following Euler-Maruyama approximation of the anchored Langevin SDE \eqref{SDE0}:
\begin{align}\label{SDE10} 
	x_{k+1}=x_k+b(x_k)(t_{k+1}-t_k)+\sqrt{2}\sigma(x_k)\left[W_{t_{k+1}}-W_{t_k}\right],
\end{align}
where $(t_k)_{k\in\mathbb{N}}$ is a partition of $[0,T]$. 
It is known that under \eqref{CN1}
and the assumptions $U_0\in C^2$ and $U\in C^1$, 
$\sup_{t\in[0,T]}\|x_k(t)-X_{t}\|\rightarrow 0$, 
in probability,
where $x_k(t)$ is the linear interpolation of $(x_k)_{k\in\mathbb{N}}$ (see e.g. \cite{GK96,G98}).
However, we are interested in controlling the discretization error uniform in time, 
and we will rely on the mean-square analysis to achieve that.
Consider the Euler-Maruyama discretization of the anchored Langevin SDE \eqref{overdamped:SDE:modified}:
\begin{equation}\label{discrete:dynamics}
	x_{k+1}=x_{k}+\eta b(x_{k})+\sqrt{2\eta}\sigma(x_{k})\xi_{k+1},
\end{equation}
where $\xi_{k+1}:=\frac{1}{\sqrt{\eta}}(W_{\eta(k+1)}-W_{\eta k})$ are i.i.d. $\mathcal{N}(0,I_{d})$ distributed and
\begin{equation}
	b(x):=-\nabla U_{0}(x)e^{U(x)-U_{0}(x)},
	\qquad
	\sigma(x):=e^{(U(x)-U_{0}(x))/2}.
\end{equation}


For $m$-strongly convex and $L$-smooth $U(x)$, 
one standard approach to obtain 2-Wasserstein convergence
is via the synchronous coupling; see e.g. \cite{DK2017}.
However, in our dynamics \eqref{discrete:dynamics},
we have a state-dependent $\sigma(x_{k})$, which
prevents us from using a straightforward  
synchronous coupling argument. 
Instead, we turn to the tool of the mean-square analysis
developed in \cite{Tao2021}, which is applicable for state-dependent 
diffusion noise as well; see e.g. \cite{TaoMirror2021}.
Let us assume the following.

\begin{assumption}\label{assump:b:sigma}
We assume that
\begin{equation}\label{discrete:condition:1}
	\langle b(x)-b(y),x-y\rangle\leq-m\Vert x-y\Vert^{2},\qquad\text{for any $x,y\in\mathbb{R}^{d}$},
\end{equation}
and 
\begin{equation}\label{discrete:condition:2}
	\Vert b(x)-b(y)\Vert\leq L\Vert x-y\Vert,\qquad\text{for any $x,y\in\mathbb{R}^{d}$},
\end{equation}
and
\begin{equation}\label{discrete:condition:3}
	\Vert\sigma(x)I_{d}-\sigma(y)I_{d}\Vert_{\mathrm{HS}}\leq\sqrt{\alpha}\Vert x-y\Vert,\qquad\text{for any $x,y\in\mathbb{R}^{d}$},
\end{equation}
where $0<\alpha<m$.
\end{assumption}

Assumption~\ref{assump:b:sigma} can be satisfied in several practical problems. For example, if $U(x)$ is strongly convex but non-smooth, 
then we can select $U_0(x)$
to be a smooth uniform approximation of $U(x)$. For example, consider regularized Bayesian regression problems with mixed $\ell_2-\ell_1$ penalty of the form
$U(x) = f(x) + m_0\|x\|^2 + \lambda \|x\|_1$
where $f(x)$ is $M$-smooth for some $M>0$ and convex. For instance, $f$ can be the least squares loss or the logistic loss. Then, $U$ is $m$-strongly convex. Let $p_\varepsilon$ be a smooth approximation of the $\ell_1$ penalty  with the property that $p_\varepsilon(x) \to p(x)$
uniformly as $\varepsilon\to 0$. For example, a common choice is $p_\varepsilon(x):= \sum_{i=1}^{d}\sqrt{x_i^2 + \varepsilon^2}$, 
with the property that $p_\varepsilon(x)\geq \|x\|_1 \geq 0 $ and $p_\varepsilon(x)-\|x\|_1 \to 0 $ uniformly on $\mathbb{R}^d$. 
In this case, 
$U_0(x)=f(x)+m_0\|x\|^2 + \lambda p_\varepsilon(x)$,
\begin{equation}\label{grad:special:case}
\nabla U_0(x)=\nabla f(x)+2m_0x
+\lambda\left(\frac{x_{1}}{\sqrt{x_{1}^{2}+\varepsilon^{2}}},\ldots,\frac{x_{d}}{\sqrt{x_{d}^{2}+\varepsilon^{2}}}\right)^{\top},
\end{equation}
and
$U(x)-U_0(x)$ admits uniformly bounded subgradients (in fact it is differentiable except when $x=0$). Furthermore, if $L_\varepsilon$ is the uniform bound for the subgradients, $L_\varepsilon\to 0$ as $\varepsilon\to 0$. Therefore, it is $L_\varepsilon$-Lipschitz. Consequently, 
\eqref{discrete:condition:3} holds if $\varepsilon$ is small enough. Similarly, \eqref{discrete:condition:1} and \eqref{discrete:condition:2} hold when $\varepsilon$ is properly chosen to be sufficiently small.  

Assumption~\ref{assump:b:sigma} can also be satisfied for sampling heavy-tailed distributions as the following results show (Corollary~\ref{cor:heavy:tailed:example:2}, Example~\ref{ex:student:discrete}).
 

%
%
\begin{corollary}\label{cor:heavy:tailed:example:2}
Let $q:\R^d\to[1,\infty)$ be a $C^1$-function such that for some $c_0,c_1,c_2>0$ and all $x,y\in\R^d$,
$$
\left\|\nabla q(x)\right\|\leq c_0\sqrt{q(x)},\ \ \left\|\nabla q(x)-\nabla q(x)\right\|\leq c_1\|x-y\|,
$$
and
$$
\<\nabla q(x)-\nabla q(x),x-y\>\geq c_2\|x-y\|^2.
$$ 
Let $\beta>dc_0^2/(4c_2)$ such that $c_3:=\int_{\R^d}q(x)^{-1-\beta}dx<\infty$. Let us choose 
\begin{equation*}
U_0(x):=\beta\log q(x),
\quad\text{and}\quad 
U(x):=(\beta+1)\log q(x).
\end{equation*}
Then Assumption~\ref{assump:b:sigma} is satisfied and
the anchored Langevin dynamics \eqref{overdamped:SDE:modified} has the unique stationary distribution 
$q(x)^{-1-\beta}/c_3$. 
\end{corollary}

\begin{example}\label{ex:student:discrete}
    Consider the $d$-dimensional Student--$t$ distribution with $\nu>0$ degrees of freedom, with mean $\mu$ and scale matrix $\Sigma\succ 0$: \[
\pi(x)\;\propto\;\left(1+\tfrac{1}{\nu}(x-\mu)^\top \Sigma^{-1}(x-\mu)\right)^{-(d+\nu)/2}.
\]
This corresponds to the potential 
$$ U(x) =\left(\tfrac{d+\nu}{2}\right)\,\log q(x),\qquad 
q(x)=1+\tfrac{1}{\nu}(x-\mu)^\top \Sigma^{-1}(x-\mu). $$
We take the anchoring potential as
$U_0(x)=\beta\log q(x)$ with $\beta=\tfrac{d+\nu}{2}-1$.
The function $q(x)$ is a strongly convex quadratic, and it satisfies Corollary \ref{cor:heavy:tailed:example:2} with 
$$c_0 = 2\sqrt{\lambda_{\max}(\Sigma^{-1})/\nu}, \quad 
c_1 = 2\lambda_{\max}(\Sigma^{-1})/\nu, \quad 
c_2 =2 \lambda_{\min}(\Sigma^{-1})/\nu.$$
Therefore, if $$\beta =\tfrac{d+\nu}{2}-1 > dc_0^2/(4c_2) = \frac{d}{2}\kappa(\Sigma) 
\quad \mbox{where} \quad \kappa(\Sigma)=\frac{\lambda_{\max}(\Sigma^{-1})}{ \lambda_{\min}(\Sigma^{-1})},$$
or equivalently if $d+\nu > 2+d\kappa(\Sigma)$
then Corollary~\ref{cor:heavy:tailed:example:2} is applicable. 
\end{example}

Next, we provide the following non-asymptotic 
result that bounds the 2-Wasserstein distance
between the distribution of the iterates $x_{k}$
in \eqref{discrete:dynamics} and the target distribution $\pi$.

\begin{theorem}\label{thm:discrete}
	Let $\nu_{k}$ denote the distribution of the iterates $x_{k}$ in \eqref{discrete:dynamics}.
	For any $0<\eta\leq\eta_{\max}$, we have
	\begin{equation}
		\mathbb{E}\Vert X_{\eta k}-x_{k}\Vert^{2}\leq C^{2}\eta,
	\end{equation}
	and
	\begin{equation}
		\mathcal{W}_{2}(\nu_{k},\pi)\leq\sqrt{2}e^{-(m-\alpha) k\eta}\mathcal{W}_{2}(\nu_{0},\pi)+\sqrt{2}C\eta^{\frac{1}{2}},
	\end{equation}
	where
	\begin{align} 
		\eta_{\max}&:=\min\Bigg\{\frac{1}{L^{2}+4\alpha},\frac{1}{4(m-\alpha)},\left(\frac{\sqrt{m-\alpha}}{20\sqrt{2}(1+4\alpha)}\right)^{2},\nonumber
  \\
  &\qquad\qquad\qquad\qquad\qquad
		\left(\frac{m-\alpha}{8\sqrt{2}(2L\sqrt{1+4\alpha}+20L(1+4\alpha))}\right)^{2}\Bigg\},
	\end{align}
	and
	\begin{equation}
		C:=\frac{2C_{1}+8LC_{2}+2\sqrt{2}C_{3}(2L\sqrt{1+4\alpha}+20L(1+4\alpha))}{m-\alpha}+\frac{2C_{2}+10\sqrt{2}(1+4\alpha)C_{3}}{\sqrt{m-\alpha}},
	\end{equation}
	where 
 \begin{align}
 &C_{1}:=3L\sqrt{1+4\alpha}\left(\Vert x_{\ast}\Vert+\Vert\sigma(x_{\ast})I_{d}\Vert_{\mathrm{HS}}\right),
\\
&C_{2}:=7(1+4\alpha)\left(\Vert x_{\ast}\Vert+\Vert\sigma(x_{\ast})I_{d}\Vert_{\mathrm{HS}}\right),
\\
&C_{3}:=\sqrt{4\mathbb{E}\Vert X_{0}\Vert^{2}+6\mathbb{E}_{X\sim\pi}\Vert X\Vert^{2}},
\end{align}
where $x_*$ is the minimizer of $U_0$.
\end{theorem}


\section{Non-Smooth Sampling}

For the overdamped Langevin SDE:
\begin{equation}
	dX_{t}=-\nabla U(X_{t})dt+\sqrt{2}dW_{t},
\end{equation}
where $U$ is non-differentiable, the most common approach
in the literature is to borrow ideas from the optimization literature
and use the subgradient or proximal method.

In our case, by considering
\begin{equation}
	dX_{t}=-\nabla U_{0}(X_{t})e^{U(X_{t})-U_{0}(X_{t})}dt
	+\sqrt{2}e^{(U(X_{t})-U_{0}(X_{t}))/2}dW_{t},
\end{equation}
where $U_{0}:\mathbb{R}^{d}\rightarrow\mathbb{R}$
plays the role as a reference potential, 
we can simply choose $U_{0}$ to be differentiable
to do the non-smooth sampling even when $U$ is non-differentiable.

In addition, in the literature of proximal Langevin methods, 
often one can write $U=f+g$ in the composition form,
where $f$ is smooth whereas $g$ is non-smooth. 
By our theory, we can choose $U_{0}=f+g_{0}$,
where $g_{0}$ is smooth, such that
\begin{equation}
	dX_{t}=-(\nabla f(X_{t})+\nabla g_{0}(X_{t}))e^{g(X_{t})-g_{0}(X_{t})}dt
	+\sqrt{2}e^{(g(X_{t})-g_{0}(X_{t}))/2}dW_{t},
\end{equation}
preserves the Gibbs distribution $\pi\propto e^{-f(x)-g(x)}$
as an invariant distribution. 
\subsection{Random time change for discretization}\label{discrete:time:change:section}

The discretization of the anchored Langevin dynamics was shown in Eq.~\eqref{discrete:dynamics}. We also proposed a random time change version of this dynamics in Section~\ref{time:change:section}. For the time change Langevin dynamics, we sample $X_t = Z_{\ell(t)}$, where $Z_t$ follows the SDE \eqref{SDE1} and $\ell(t)$ is determined by the ODE:
\begin{equation}
	d\ell(t)=\exp\left\{U(Z_{\ell(t)})-U_0(Z_{\ell(t)})\right\}dt.
\end{equation}
Let $z$ be the discretized variable of $Z$, then at every iteration $k$, we have access to $\ell_k$ and $z_{\ell_k}$, where we denote $\ell_k$ as the discretized $\ell(t)$ and  $z_{\ell_k}$ as the value of $z$ at iteration $k$ with the following scheme.
First we update the random time change $\ell$ by:
\begin{equation}
	\ell_{k+1}=\ell_k+\eta\exp\left\{U(z_{\ell_k})-U_0(z_{\ell_k})\right\},
\end{equation}
and next, we update $z$ as:
\begin{equation}\label{time-change}
	z_{\ell_{k+1}} = z_{\ell_k} - \Delta \ell_k\nabla U_0(z_{\ell_k})+\sqrt{2\Delta \ell_k}\xi_{k+1},
\end{equation}
where we use $\Delta \ell_k = \ell_{k+1}-\ell_k$ as the stepsize. Then we set $x_{k+1} = z_{\ell_{k+1}}$ to be the updated $x$ at iteration $k+1$. Hence, for $x_0 = z_{\ell_0}$, we have $x_k = z_{\ell_k}$ for any $k$.

\begin{theorem}\label{thm:equivalence}
	Assume synchronous coupling between the anchored Langevin dynamics and the random time change Langevin dynamics and fix an initial $x_0$. The discretizations of the two algorithms are equivalent.
\end{theorem}

Hence, by Theorem~\ref{thm:equivalence}, we will only show the performance of the anchored Langevin dynamics in comparison to the original Langevin dynamics with reference potential $U_0$.

\subsection{Gaussian smoothing algorithms}\label{section:smooth}

We consider the case when the target function $U(x)$
is not differentiable and hence we do not have
access to $\nabla U(x)$ and overdamped Langevin algorithm is not directly applicable. 
When the target is not differentiable, smoothing methods have been studied in the literature, 
including the Gaussian smoothing for LMC algorithms (see e.g. \cite{Chatterji20} and \cite{nesterov2017random}). 
We consider the case that the target function $U(x)$ can be written as the sum
of a smooth function $f(x)$ and a non-smooth function $g(x)$:
\begin{equation}
U(x)=f(x)+g(x).
\end{equation}
We define $U_{0}(x)$ as
\begin{equation}\label{defn:U:0:f:g:0}
U_{0}(x)=f(x)+g_{0}(x),
\end{equation}
where $g_{0}(x)$ is the Gaussian smoothing function of $g(x)$:
\begin{equation}\label{defn:g:0}
g_{0}(x)=\mathbb{E}_{\xi}[g(x+\mu\xi)],\qquad\xi\sim\mathcal{N}(0,I_{d}).
\end{equation}

We next assume that $g(x)$ is Lipschitz continuous. We note that by Rademacher's theorem, Lipschitz functions are almost everywhere differentiable. For such functions, the Clarke subdifferential $\partial G(x)$ at every point $x\in\mathbb{R}^d$ exists and is a compact set; see \cite{rockafellar2009variational,qi1993nonsmooth} for the definition of the Clarke subdifferential.

\begin{assumption}\label{assump:Lip:g}
Assume that $g(x)$ is $K$-Lipschitz, i.e.
\begin{equation}
|g(x)-g(y)|\leq K\Vert x-y\Vert,\qquad\text{for any $x,y\in\mathbb{R}^{d}$},
\end{equation}
Furthermore, assume that $g$ is $\gamma$-weakly convex for some $\gamma\geq 0$, i.e. the function $g_\gamma(x):= g(x) + \frac{\gamma}{2}\|x\|^2$ is convex.
\end{assumption}

A well-known property of $\gamma$-weakly convex functions is that they satisfy
$$ 
g(y) \geq g(x) + \langle u, y-x \rangle -\frac{\gamma}{2}\|y-x\|^2, 
$$
(see e.g. \cite{davis2018subgradient}). Such functions arise frequently in applications including deep learning, logistic regression and data science (see e.g. \cite{zhu2023distributionally,zhangsapdplus,davis2018subgradient}). 
For example, when $g(x)=\|x\|_1$, Assumption~\ref{assump:Lip:g} is satisfied.
Similarly, the SCAD and MCP penalties discussed in Section~\ref{experiments} are piecewise twice continuously differentiable admitting directional derivatives and satisfy Assumption~\ref{assump:Lip:g}. In addition, we will assume that $f(x)$ is smooth and strongly convex.

\begin{assumption}\label{assump:smooth:f}
Assume that $f(x)$ is differentiable and $L_{f}$-smooth, i.e.
\begin{equation}
\Vert\nabla f(x)-\nabla f(y)\Vert\leq L_{f}\Vert x-y\Vert,\qquad\text{for any $x,y\in\mathbb{R}^{d}$},
\end{equation}
and further assume that $f(x)$ is $m_{f}$-strongly convex, i.e.
\begin{align*}
\left\langle\nabla f(x)-\nabla f(y),x-y\right\rangle\geq m_{f}\Vert x-y\Vert^{2},\qquad\text{for any $x,y\in\mathbb{R}^{d}$}.
\end{align*}
\end{assumption}

First, we will show that under Assumption~\ref{assump:Lip:g}, 
the difference between $U$ and $U_{0}$ is uniformly bounded.

\begin{lemma}\label{lem:CP:bound}
    Under Assumption~\ref{assump:Lip:g}, we have
    $$
    |U(x)-U_0(x)|\leq K\mu\sqrt{d}.
    $$
\end{lemma}

\begin{lemma}\label{ineq-small-grad-error} Let Assumption~\ref{assump:Lip:g}  hold. Then we have
\begin{equation} 
\sup_{x\in\mathbb{R}^d} \mbox{dist}\left(\nabla g_0(x), \partial g(x)\right)
\leq 3^{3/4}C_{g} \mu d^{3/2}, 
\end{equation}
where $\mbox{dist}(\cdot, \partial g(x))$ denotes the Hausdorff distance to the set $\partial g(x)$.
\end{lemma}

\begin{proposition}\label{prop-when-assump-hold}
Suppose Assumptions~\ref{assump:Lip:g} and \ref{assump:smooth:f} hold. Let $U(x) = f(x) + m_0\|x\|^2 + g(x)$ and assume $U_0= f(x) + m_0\|x\|^2 + g_0(x)$ is $c_0$-strongly convex and $L_0$-smooth with $c_0 = \Theta(1)$ as $\mu \to 0$. 
Assume also
$$ \sup_{x\in\mathbb{R}^{d}} \left\|\nabla U_0(x)\left(e^{g(x)-g_0(x)}-1\right)\right\| = o(\mu),$$
$$ \sup_{x\in\mathbb{R}^{d}} \left\|U_0(x) \cdot e^{g(x)-g_0(x)}\cdot (\partial g(x)-\nabla g_0(x))\right\| = o(\mu),$$ 
as $\mu\rightarrow 0$. Then, for $\mu$ small enough, Assumption~\ref{assump:b:sigma} holds. 
\end{proposition}

\begin{remark}
When $g(x) = \|x\|_1$, under the Gaussian smoothing, we have $g_0(x) = \mathbb{E} \|x+\mu\xi\|_1=\sum_{i=1}^{d}\mathbb{E} |x_{i}+\mu\xi_{i}|$, where $\xi=(\xi_{1},\ldots,\xi_{d})\sim\mathcal{N}(0,I_{d})$. 
Here, by noticing that for every $i$, $|x_{i}+\mu\xi_{i}|$
is a folded normal distribution \cite{Leone1961}, we can compute that
    $$ g_0(x) = \sum_{i=1}^{d}\mathbb{E} |x_{i}+\mu\xi_{i}| = 
  \sum_{i=1}^d \left( \mu \frac{\sqrt{2}}{\sqrt{\pi}} e^{-\frac{x_i^2}{2\mu^2}} + x_i \left(1-2\Phi\left(-\frac{x_i}{\mu}\right) \right)\right), $$
  where
  ${\displaystyle \Phi (x)\;=\;{\frac {1}{2}}\left[1+\mathrm{erf} \left({\frac {x}{\sqrt {2}}}\right)\right]}$ 
  is the normal cumulative density function with $\mbox{erf}(x)=\frac{2}{\sqrt{\pi}}\int_{t=0}^x e^{-t^2}dt$. For simplicity, assume $f(x) = 0$ and $m_0>0$. Then, for any $\mu>0$, 
  \begin{align*} \nabla_{x_i} U_0(x) &=  m_0 x_i + \left(-\frac{x_i}{\mu}  \frac{\sqrt{2}}{\sqrt{\pi}} e^{-\frac{x_i^2}{2\mu^2}} +  \left(1-2\Phi\left(-\frac{x_i}{\mu}\right) \right)  
  +\frac{2x_i}{\mu} \Phi'\left(-\frac{x_i}{\mu}\right)
  \right) \\
  &=m_0 x_i + \left(-\frac{x_i}{\mu}  \frac{\sqrt{2}}{\sqrt{\pi}} e^{-\frac{x_i^2}{2\mu^2}} +  \left(1-2\Phi\left(-\frac{x_i}{\mu}\right) \right)  
  +\frac{2x_i}{\mu} \frac{1}{\sqrt{2\pi}}e^{-\frac{x_i^2}{2\mu^2}}\right)
  \\
  &=m_{0}x_i+1-2\Phi\left(-\frac{x_i}{\mu}\right) = m_{0}x_i-\mathrm{erf}\left(-\frac{x_i}{\mu\sqrt{2}}\right),
  \end{align*}
and in addition,  
 $$ e^{g(x)-g_0(x)}-1 = \left( \prod_{i=1}^d  e^{|x_i| -\mu \frac{\sqrt{2}}{\sqrt{\pi}} e^{-\frac{x_i^2}{2\mu^2}} - x_i \left(1-2\Phi\left(-\frac{x_i}{\mu}\right) \right) }\right)  -1 .$$

Since the function $g(x)-g_0(x)$ is even, we have
\begin{align*} 
\sup_{x\in\mathbb{R}^{d}} \left|e^{g(x)-g_0(x)}-1\right| &= \sup_{x_i\geq 0, \forall i} \Bigg|\left( \prod_{i=1}^d  e^{|x_i| -\mu \frac{\sqrt{2}}{\sqrt{\pi}} e^{-\frac{x_i^2}{2\mu^2}} - x_i \left(1-2\Phi\left(-\frac{x_i}{\mu}\right) \right) }\right)  -1\Bigg|\\
 &= \sup_{x_i\geq 0, \forall i} \Bigg|\left( \prod_{i=1}^d  e^{-\mu \frac{\sqrt{2}}{\sqrt{\pi}} e^{-\frac{x_i^2}{2\mu^2}} + x_i 2\Phi\left(-\frac{x_i}{\mu}\right)  }\right)  -1\Bigg| \\
 &\leq  \sup_{x_i\geq 0, \forall i} \left( \prod_{i=1}^d  e^{-\mu \frac{\sqrt{2}}{\sqrt{\pi}} e^{-\frac{x_i^2}{2\mu^2}} + x_i 2\Phi\left(-\frac{x_i}{\mu}\right)  }\right) -1.
\end{align*}
Note that as $x_i\to\infty$,  $x_i 2\Phi\left(-\frac{x_i}{\mu}\right)\to 0$. Therefore, $r(x_i):= -\mu \frac{\sqrt{2}}{\sqrt{\pi}} e^{-\frac{x_i^2}{2\mu^2}} + x_i 2\Phi\left(-\frac{x_i}{\mu}\right)  \to 0$ when $x_i \to \infty$. Similarly, $r(x_i)\to 0$ when $x_i\to -\infty$ or when $x_i\to 0$. Therefore, we can argue that $\sup_{x\in\mathbb{R}^{d}} |e^{g(x)-g_0(x)}-1| = o(\mu)$. Similarly, after straightforward computations, it can be shown that
$$\sup_{x\in\mathbb{R}^{d}}\left|\nabla_{x_i} U_0(x) \left(e^{g(x)-g_0(x)}-1\right) \right| \leq \sup_{x_i\geq 0}\left| m_{0}x_i+\mathrm{erf}\left(-\frac{x_i}{\mu\sqrt{2}}\right)\right|
\cdot \sup_{x\in\mathbb{R}^{d}} \left|e^{g(x)-g_0(x)}-1\right| = o(\mu),
$$
and
$\sup_{x\in\mathbb{R}^{d}} \left\|U_0(x) \cdot e^{g(x)-g_0(x)}\cdot (\partial g(x)-\nabla g_0(x))\right\| = o(\mu)$. Then, from Proposition~\ref{prop-when-assump-hold}, we conclude that Assumption~\ref{assump:b:sigma} holds.
\end{remark}

Under Assumption~\ref{assump:Lip:g}, $g$ is continuous.
Denote $z = x + \mu\xi$. By applying Leibniz integral rule, we can compute that
\begin{align}
    \nabla_{x}g_0(x) &= \nabla_{x}\mathbb{E}_{\xi}[g(x+\mu\xi)] 
=\nabla_{x}\left(\int_{\mathbb{R}^{d}}g(x+\mu\xi)\frac{1}{(2\pi)^{d/2}}e^{-\frac{\Vert\xi\Vert^{2}}{2}}d\xi\right)
    \nonumber\\
&=\nabla_{x}\left(\int_{\mathbb{R}^{d}}\frac{g(z)}{\mu^{d}}\frac{1}{(2\pi)^{d/2}}e^{-\frac{\Vert z-x\Vert^{2}}{2\mu^{2}}}dz\right)=\int_{\mathbb{R}^{d}}\frac{g(z)}{\mu^{d}}\frac{(z-x)}{\mu^2}\frac{1}{(2\pi)^{d/2}}e^{-\frac{\Vert z-x\Vert^{2}}{2\mu^{2}}}dz\nonumber\\
    &=\frac{1}{\mu}\int_{\mathbb{R}^{d}}g(x+\mu\xi)\cdot\xi\cdot \frac{1}{(2\pi)^{d/2}}e^{-\frac{\Vert\xi\Vert^{2}}{2}}d\xi
    = \frac{1}{\mu}\mathbb{E}_{\hat{\xi}}\left[g\left(x+\mu\hat{\xi}\right)\hat{\xi}\right],\label{eq-grad-smoothed}
\end{align}
where $\hat{\xi}\sim\mathcal{N}(0,I_{d})$. Hence, using smoothing functions $g_0(x)$, we no longer need access to the gradients of $g(x)$. Since the anchored Langevin dynamics requires access to the values of $U_0(x)$ and $\nabla U_0(x)$, we will use Monte Carlo simulations to approximate these expectations, where the simulations are independent. In practice, we can approximate $U_{0}(x_{k})$ by
\begin{equation}
    \tilde{U}_{0}(x_{k})
    :=f(x_{k})+\frac{1}{N}\sum_{i=1}^{N}g\left(x_{k}+\mu\xi_{i,k}\right),
\end{equation}
and approximate $\nabla U_{0}(x_{k})$
by
\begin{equation}
    \nabla\tilde{U}_{0}(x_{k})
    :=\nabla f(x_{k})+\frac{1}{\mu N}\sum_{i=1}^{N}\hat{\xi}_{i,k}g\left(x_{k}+\mu\hat{\xi}_{i,k}\right),
\end{equation}
where $\xi_{i,k}$'s and $\hat{\xi}_{i,k}$'s are i.i.d. $\mathcal{N}(0,I_{d})$.
We then obtain Algorithm~\ref{anchored:algo} for the anchored Langevin dynamics.

\begin{algorithm}[ht]
    \caption{Anchored Langevin dynamics with Gaussian smoothing}\label{anchored:algo}
    \begin{algorithmic}
        \Require $n, N, \eta > 0$, $\mu$, $U(x)=f(x)+g(x)$
        \State Initialize a random $x_0$;
        \For{$k \gets 1$ to $n$}
        \State Approximate $U_0(x_k)$ by $\tilde{U}_0(x_k) =f(x_k)+ \frac{1}{N}\sum_{i=1}^{N}g(x_k+\mu\xi_{i,k})$ for $\xi_{i,k}\sim \mathcal{N}(0,I_{d})\,\forall i$;
        \State Approximate $\nabla U_0(x_k)$ by $\nabla \tilde{U}_0(x_k) =\nabla f(x_{k})+ \frac{1}{\mu N}\sum_{i=1}^{N}\hat{\xi}_{i,k}g(x_k+\mu\hat{\xi}_{i,k})$ for $\hat{\xi}_{i,k}\sim \mathcal{N}(0,I_{d})\,\forall i$;
        \State Compute $x_{k+1}$ using the Euler-Maruyama discretization in Eq.~\eqref{discrete:dynamics}:
        \begin{equation*}
            x_{k+1} \gets x_{k}-\eta \nabla \tilde{U}_{0}(x_k)e^{U(x_k)-\tilde{U}_{0}(x_k)}+\sqrt{2\eta}e^{(U(x_k)-\tilde{U}_{0}(x_k))/2}\xi_{k+1}\text{ for }\xi_{k+1}\sim \mathcal{N}(0,I_{d});
        \end{equation*}
        \EndFor
    \end{algorithmic}
\end{algorithm}

On the other hand, for the random time change Langevin dynamics, we use the discretization scheme in Section~\ref{discrete:time:change:section} to get Algorithm~\ref{time:change:algo}.

\begin{algorithm}[ht] 
    \caption{Random time change Langevin dynamics with Gaussian smoothing}\label{time:change:algo}
    \begin{algorithmic}
        \Require $n, N, \eta > 0$, $\ell_0=0$, $\mu$, $U(x)=f(x)+g(x)$
        \State Initialize a random $x_0$;
        \State Set $z_{\ell_0} \gets x_0$
        \For{$k \gets 1$ to $n$}
        \State Approximate $U_0(z_{\ell_k})$ by $\tilde{U}_0(z_{\ell_k}) =f(z_{\ell_k})+ \frac{1}{N}\sum_{i=1}^{N}g(z_{\ell_k}+\mu\xi_{i,k})$ for $\xi_{i,k}\sim \mathcal{N}(0,I_{d})\,\forall i$;
        \State Approximate $\nabla U_0(z_{\ell_k})$ by $\nabla \tilde{U}_0(z_{\ell_k}) = \nabla f(z_{\ell_k})+\frac{1}{\mu N}\sum_{i=1}^{N}\hat{\xi}_{i,k}U(z_{\ell_k}+\mu\hat{\xi}_{i,k})$ for $\hat{\xi}_{i,k}\sim \mathcal{N}(0,I_{d})\,\forall i$;
        \State $\ell_{k+1} \gets \ell_k+\eta\exp\left\{U(z_{\ell_k})-\tilde{U}_0(z_{\ell_k})\right\}$;
        \State $z_{\ell_{k+1}} \gets z_{\ell_k} - \Delta \ell_k\nabla \tilde{U}_0(z_{\ell_k})+\sqrt{2\Delta \ell_k}\xi_{k+1}$ for $\Delta \ell_k = \ell_{k+1}-\ell_k$ and $\xi_{k+1}\sim \mathcal{N}(0,I_{d})$;
        \State $x_{k+1} \gets z_{\ell_{k+1}}$;
        \EndFor
    \end{algorithmic}
\end{algorithm}

\begin{remark}
    By Theorem~\ref{thm:equivalence}, Algorithms~\ref{anchored:algo} and \ref{time:change:algo} are equivalent.
\end{remark}

\begin{remark}
    It can be shown that $U_0$ is smooth even if $U$ is not and $U_0$ preserves the strong convexity of $U$.
\end{remark}

\subsection{Non-asymptotic analysis for Gaussian smoothing algorithms}

We obtain the following anchored Langevin dynamics
with Gaussian smoothing
that can be used to sample a target
distribution whose density
is not necessarily differentiable:
\begin{equation}\label{discrete:dynamics:CP:smoothing}
    \tilde{x}_{k+1}=\tilde{x}_{k}+\eta \tilde{b}(\tilde{x}_{k})+\sqrt{2\eta}\tilde{\sigma}(\tilde{x}_{k})\xi_{k+1},
\end{equation}
where $\xi_{k+1}:=\frac{1}{\sqrt{\eta}}(W_{\eta(k+1)}-W_{\eta k})$ are i.i.d. $\mathcal{N}(0,I_{d})$ distributed and
\begin{equation}
    \tilde{b}(x):=-\nabla \tilde{U}_{0}(x)e^{U(x)-\tilde{U}_{0}(x)},
    \qquad
    \tilde{\sigma}(x):=e^{(U(x)-\tilde{U}_{0}(x))/2}.
\end{equation}

Let $\tilde{\nu}_{k}$ denote the distribution
of the iterates $\tilde{x}_{k}$. 
We aim to derive an explicit upper bound on:
$\mathcal{W}_{2}(\tilde{\nu}_{k},\pi)$.
Starting at a common point $x_0 \sim \nu_0$, consider the Euler-Maruyama discretization of the anchored Langevin SDE and the anchored Langevin dynamics with Gaussian smoothing at step $k \in \mathbb{N}^*$ with synchronous coupling as follows:
\begin{align}
    x_{k+1} &=x_{k}+\eta b(x_{k})+\sqrt{2\eta}\sigma(x_{k})\xi_{k+1},\label{discrete:overdamped:k} \\
    \tilde{x}_{k+1} &=\tilde{x}_{k}+\eta\tilde{b}(\tilde{x}_{k})+\sqrt{2\eta}\tilde{\sigma}(\tilde{x}_{k})\xi_{k+1}\label{discrete:MC:k},
\end{align}
where $\xi_{k+1}$'s are i.i.d. $\mathcal{N}(0,I_{d})$ distributed. The following expectations are conditional on $x$, and thus for simplicity, we will write $\mathbb{E}[\bullet]$ instead of $\mathbb{E}[\bullet|x]$. 
We will show that in the 2-Wasserstein distance, 
$\tilde{x}_{k}$ is close to $x_{k}$. 

Next, we recall from \eqref{defn:U:0:f:g:0} and \eqref{defn:g:0} that $U_{0}(x)=f(x)+g_{0}(x)$, where $g_0(x):=\mathbb{E}[g(x+\mu\xi)]$,
with $\xi\sim\mathcal{N}(0,I_{d})$ 
and we assume that Assumption~\ref{assump:Lip:g} holds.
We will show that $\tilde{b}$ is close to $b$ and $\tilde{\sigma}$ is close to $\sigma$. 
First, we present the following technical lemma.

\begin{lemma}\label{lem:exposquared:CP}
    For $x \in \mathbb{R}^d$, we have the following inequality:
    \begin{align}
        \mathbb{E}\left[\left| e^{U(x)-\tilde{U}_{0}(x)}-e^{U(x)-U_{0}(x)}\right|^2\right] &\leq \frac{4K\mu\sqrt{d}}{\sqrt{N}}\cdot  e^{6K\mu\sqrt{d}}. \label{lem:exposquared:2:CP}
    \end{align}
\end{lemma}

\begin{remark}
    One can check that for $0 \leq x \leq 1$, we have $e^x \leq 1+ex$. Since we will generally choose $\mu$ to be small, we can choose $\mu$ such that $6K\mu\sqrt{d} \leq 1$, i.e. $\mu\leq\frac{1}{6K\sqrt{d}}$, then the inequality in Lemma~\ref{lem:exposquared:CP} becomes:
    \begin{align*}
        \mathbb{E}\left[\left| e^{U(x)-\tilde{U}_{0}(x)}-e^{U(x)-U_{0}(x)}\right|^2\right] \leq \frac{4K\mu\sqrt{d}}{\sqrt{N}}\left(1+6eK\mu\sqrt{d}\right) \leq \frac{1}{\sqrt{N}}(1+e).
    \end{align*}
\end{remark}


Now, we are ready to show that $\tilde{b}$ is close to $b$ and $\tilde{\sigma}$ is close to $\sigma$. 

\begin{theorem}\label{thm:MC:CP}
    For all $x \in \mathbb{R}^d$ and $\mu\leq 1/(6K\sqrt{d})$, we have the following results for the Monte Carlo approximation $\tilde{U}_0(x)$ and $\nabla\tilde{U}_0(x)$:
    \begin{align}
        &\mathbb{E}\left[\Vert\tilde{b}(x)-b(x)\Vert^2\right] \leq \frac{1}{\mu^2\sqrt{N}}(A_{1}\Vert x\Vert^{2}+A_{2}), \label{bound:b:CP}\\
        &\mathbb{E}\left[\left|\tilde{\sigma}(x)-\sigma(x)\right|^2\right] \leq \frac{1}{\sqrt{N}}B, \label{bound:sigma:CP}
    \end{align}
    where
\begin{align}
&A_{1}:=4\left(1+e\right)\left(2\mu^{2}L_{f}^{2}+8K^{2}d\right),\label{eqn:defn:C:1}
        \\
        &A_{2}:= 4\left(1+e\right)\left(2\mu^{2}L_{f}^{2}\Vert x_{\ast}^{f}\Vert^{2}
+13\mu^{2}K^{2}d^{2}+8(g(0))^{2}d\right),\label{eqn:defn:C:2}
        \\
        &B:=\frac{1}{3}\left(1+\frac{1}{2}e\right),\label{A:B:x:CP}
\end{align}
where $x_{\ast}^{f}$ is the unique minimizer of $f$.
\end{theorem}


We recall that
\begin{equation}
\tilde{x}_{k+1} =\tilde{x}_{k}+\eta\tilde{b}(\tilde{x}_{k})+\sqrt{2\eta}\tilde{\sigma}(\tilde{x}_{k})\xi_{k+1}.
\end{equation}
Next, we provide a uniform $L^{2}$ bound for $\tilde{x}_{k}$.

\begin{lemma}\label{lem:L2}
Assume $\eta\leq\frac{m\mu^{2}}{4e^{6K\mu\sqrt{\mu}d}(4\mu^{2}L_{f}^{2}+8K^{2}d)}$
and 
$N\geq\left(\frac{4\sqrt{2A_{1}}}{m\mu}\right)^{4}$.
For any $k\in\mathbb{N}$,
\begin{align*}
\mathbb{E}\Vert\tilde{x}_{k}\Vert^{2} 
&\leq
2\Vert x_{\ast}\Vert^{2}+
2\mathbb{E}\Vert\tilde{x}_{0}-x_{\ast}\Vert^{2}
+\frac{4}{m}e^{3K\mu\sqrt{d}}d
+\frac{4}{m}\frac{\sqrt{2A_{1}}}{\mu N^{1/4}}\left(\Vert x_{\ast}\Vert^{2}+\frac{A_{2}}{2A_{1}}\right)
\\
&+\frac{2\eta}{m}\frac{2e^{6K\mu\sqrt{d}}}{\mu^{2}}
\left((4\mu^{2}L_{f}^{2}+8K^{2}d)\Vert x_{\ast}\Vert^{2}
+2\mu^{2}L_{f}^{2}\Vert x_{\ast}^{f}\Vert^{2}
+2K^{2}\mu^{2}(3d^{2})
+4(g(0))^{2}d\right),
\end{align*}
where $x_{\ast}$ is the minimizer of $U_{0}$ and $x_{\ast}^{f}$ is the unique minimizer of $f$, 
and $A_{1},A_{2}$ are defined in \eqref{eqn:defn:C:1}-\eqref{eqn:defn:C:2}.
\end{lemma}

An immediate consequence of Theorem~\ref{thm:MC:CP} and Lemma~\ref{lem:L2}
is the following corollary.

\begin{corollary}\label{cor:tilde:b:b}
Under the assumptions of Theorem~\ref{thm:MC:CP} and Lemma~\ref{lem:L2}, for any $k\in\mathbb{N}$,
    \begin{align}
        &\mathbb{E}\left[\left\Vert\tilde{b}(\tilde{x}_{k})-b(\tilde{x}_{k})\right\Vert^2\right] \leq \frac{1}{\mu^2\sqrt{N}}A, \\
        &\mathbb{E}\left[\left|\tilde{\sigma}(\tilde{x}_{k})-\sigma(\tilde{x}_{k})\right|^2\right] \leq \frac{1}{\sqrt{N}}B, 
    \end{align}
where $B$ is defined in \eqref{A:B:x:CP} and 
\begin{align}
A&:=2A_{1}\Vert x_{\ast}\Vert^{2}
+2A_{1}\mathbb{E}\Vert\tilde{x}_{0}-x_{\ast}\Vert^{2}
+\frac{4A_{1}}{m}e^{3K\mu\sqrt{d}}d
+\frac{4A_{1}}{m}\frac{\sqrt{2A_{1}}}{\mu N^{1/4}}\left(\Vert x_{\ast}\Vert^{2}+\frac{A_{2}}{2A_{1}}\right)
\nonumber
\\
&\qquad
+\frac{2A_{1}\eta}{m}\frac{2e^{6K\mu\sqrt{d}}}{\mu^{2}}
\Big((4\mu^{2}L_{f}^{2}+8K^{2}d)\Vert x_{\ast}\Vert^{2}
+2\mu^{2}L_{f}^{2}\Vert x_{\ast}^{f}\Vert^{2}\nonumber
\\
&\qquad\qquad\qquad\qquad\qquad\qquad
+2K^{2}\mu^{2}(3d^{2})
+4(g(0))^{2}d\Big)+A_{2},\label{eqn:defn:A}
\end{align}
where $x_{\ast}$ is the minimizer of $U_{0}$ and $x_{\ast}^{f}$ is the unique minimizer of $f$, 
and $A_{1},A_{2}$ are defined in \eqref{eqn:defn:C:1}-\eqref{eqn:defn:C:2}.
\end{corollary}

We then obtain the following result:
\begin{proposition}\label{prop:MC:CP}
    For $\nu_k$ and $\tilde{\nu}_k$ being the distributions of $x_k$ and $\tilde{x}_k$ respectively, the following inequality holds:
    \begin{equation}
        \mathcal{W}_2(\nu_k,\tilde{\nu}_k) \leq\tau\left(\left(1+\eta+2\eta L^2+4\eta d\alpha+2\eta^2L^2\right)^{k}-1\right),
    \end{equation}
    where
    \begin{equation*}
        \tau := \frac{(2\eta+2)\frac{1}{\mu^2\sqrt{N}}A+4d\frac{1}{\sqrt{N}}B}{1+2\eta L^2+2L^2+4d\alpha},
    \end{equation*}
    with $A,B$ defined in \eqref{eqn:defn:A} and \eqref{A:B:x:CP}.
\end{proposition}

By combining Theorem~\ref{thm:discrete} and Proposition~\ref{prop:MC:CP}, 
we obtain the following theorem that provides the 2-Wasserstein distance
between the distribution of 
the $k$-th iterate of the anchored Langevin dynamics with Gaussian smoothing
and the Gibbs distribution.

\begin{theorem}\label{thm:MC:final:CP}
    Under Assumptions~\ref{assump:b:sigma}, \ref{assump:Lip:g} and \ref{assump:smooth:f}, for $\mu\leq 1/(6K\sqrt{d})$, 
$N\geq\left(\frac{4\sqrt{2A_{1}}}{m\mu}\right)^{4}$ and $\eta\leq\frac{m\mu^{2}}{4e^{6K\mu\sqrt{\mu}d}(4\mu^{2}L_{f}^{2}+8K^{2}d)}$, the distribution $\tilde{\nu}_k$ of the $k$-th iterate of the anchored Langevin dynamics with Gaussian smoothing satisfies the following result:
    \begin{equation}
        \mathcal{W}_2(\tilde{\nu}_k,\pi) \leq\sqrt{2}e^{-(m-\alpha)k\eta}\mathcal{W}_{2}(\nu_{0},\pi)+\sqrt{2}C\eta^{\frac{1}{2}}+\tau\left((1+\eta\varrho)^{k}-1\right),
    \end{equation}
    where $\varrho := 1+4L^2+4d\alpha$ and $C$ and $\tau$ are defined in Theorem~\ref{thm:discrete} and Proposition~\ref{prop:MC:CP}.
\end{theorem}

Given Theorem~\ref{thm:MC:final:CP}, 
we are able to show that
we can achieve $\epsilon$-accuracy for
the 2-Wasserstein distance
between the distribution of 
the $k$-th iterate of the anchored Langevin dynamics with Gaussian smoothing
and the Gibbs distribution by properly choosing $\mu,k,\eta$ and $N$.

\begin{corollary}\label{cor:final:complexity:CP}
    For $\epsilon>0$, if we choose $\mu$, $k$, $\eta$ and $N$ that satisfy:
    \begin{align*}
        &\mu \leq \frac{1}{6K\sqrt{d}},\qquad k\eta\geq \frac{1}{\beta}\log\left(\frac{2\sqrt{2}\mathcal{W}_{2}(\nu_{0},\pi)}{\epsilon}\right), \\
        &\eta \leq \min\left(\left(\frac{\epsilon}{4\sqrt{2}C}\right)^{2},\frac{m\mu^{2}}{4e^{6K\mu\sqrt{\mu}d}(4\mu^{2}L_{f}^{2}+8K^{2}d)}\right), \\
        &N \geq \max\left(\left(\frac{\left(\frac{4}{\mu^2}(2\eta+2)A+16dB\right)\left(e^{\eta k\varrho}-1\right)}{\epsilon(1+2L^2+4d\alpha)}\right)^2,\left(\frac{4\sqrt{2A_{1}}}{m\mu}\right)^{4}\right),
    \end{align*}
    then we have $\mathcal{W}_2(\tilde{\nu}_k,\pi)\leq\epsilon$.
\end{corollary}

\section{Numerical Experiments}\label{experiments}

In this section, we conduct some numerical experiments to validate our theory and investigate the performance of the anchored Langevin dynamics. We specifically target distributions whose densities are not differentiable, i.e. the gradient of the target is inaccessible at finitely many points, where the classical overdamped Langevin algorithm is not feasible.  We will apply our algorithm to simulating Laplace distributions (univariate and multivariate), Bayesian logistic regression with regularizers and feedforward neural network with ReLU activation on real data sets. We will perform our experiments using Algorithm~\ref{anchored:algo} and use the original (overdamped) Langevin dynamics with reference potential $U_0$ (see \cite{Chatterji20}) as a benchmark with the SDE:
\begin{equation}\label{original}
    dX_t = -\nabla U_0(X_t)dt+\sqrt{2}dW_t.
\end{equation}
The discretization of Eq. \eqref{original} is as follows:
\begin{equation}
    x_{k+1} = x_k - \eta\nabla U_0(x_k)+\sqrt{2\eta}\xi_{k+1},
\end{equation}
where $\eta > 0$ is the stepsize, or learning rate, and $\xi_k$ are i.i.d. random noise with the distribution $\mathcal{N}(0,I_d)$. The Wasserstein distance metric uses $\pi \propto e^{-U(x)}$ with the expectations being estimated by Monte Carlo simulations.

\subsection{Simulating Laplace distributions}

In this section, we will simulate the univariate and multivariate Laplace distributions. Since Laplace distributions have non-differentiable points, the conventional gradient descent algorithm will not work without some control assumptions. Hence, we will show that our algorithm using Gaussian smoothing as the reference potential can overcome this limitation and converge nicely.

\subsubsection{Univariate Laplace distribution}

Univariate Laplace distribution has the following p.d.f.:
$\pi(x;\alpha,b) = \frac{1}{2b} \exp\left(-\frac{|x-\alpha |}{b}\right)$.
We will simulate $5,000$ data points and estimate the 2-Wasserstein distance between the simulated distribution and the true univariate Laplace distribution. 
The one-dimensional 2-Wasserstein can be estimated as:
$\mathcal{W}_2(X,\mathcal{L}) = \sqrt{\frac{1}{n}\sum_{i=1}^{n}(X_i-Q_i)^2}$,
where $n=5000$ is the sample size, $X_i$ is the $i$-th data point of the sorted sample and $Q_i$ is the $(i/n)$-th quantile of the Laplace distribution. Since the quantiles at the area of the two tails are too close to positive or negative infinity, we will ignore the 1\% tail on each side of the distribution and measure the 2-Wasserstein distance using the middle 98\%.
\begin{figure}
    \begin{subfigure}{.5\textwidth}
        \centering
        \includegraphics[width=1\linewidth, height=0.7\linewidth]{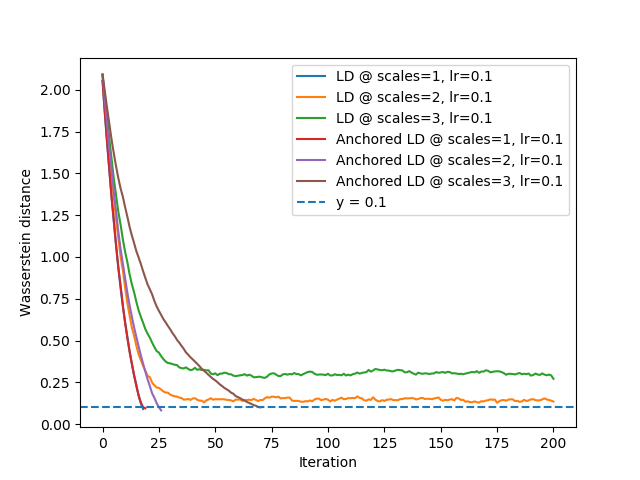}
        \caption{Prior $\mathcal{N}(0,10)$}
        \label{1dLaplacea}
    \end{subfigure}
    \begin{subfigure}{.5\textwidth}
        \centering
        \includegraphics[width=1\linewidth, height=0.7\linewidth]{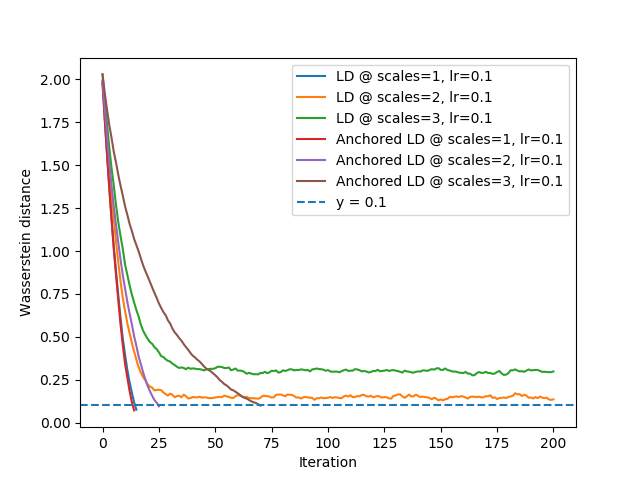}
        \caption{Prior Uniform$(-5,5)$}
        \label{1dLaplaceb}
    \end{subfigure}
    \caption{Performance of the Langevin algorithms with Gaussian smoothing reference on simulating univariate Laplace distribution $\pi(x) \propto \exp (-\sqrt{2}|x|)$.}
    \label{1dLaplace}
\end{figure}
We choose the following hyper-parameters: $U(x) \propto \sqrt{2}|x|$ (corresponding to $\alpha = 0$, $b = \frac{1}{\sqrt{2}}$), each of the expectations is estimated by $500$ Monte Carlo simulations and the initial distribution is $\mathcal{N}(0,10I)$. The results of the models are shown in Figure~\ref{1dLaplace} with three different levels of noise and stepsize $\eta=0.1$. From two different prior distributions, both Figures~\ref{1dLaplacea} and \ref{1dLaplaceb} show that on average, anchored Langevin algorithm can achieve lower Wasserstein distance compared to the original Langevin dynamics.

\subsubsection{Multivariate Laplace distribution}

Symmetric multivariate Laplace distribution has the characteristic function (see e.g. \cite{Kotz01}):
\begin{equation}\label{d-laplace}
    \Phi(t;m,\Sigma) = \frac{\exp(im^{\top}t)}{1+\frac{1}{2}t^{\top}\Sigma t},
\end{equation}
where $m$ is the mean vector and $\Sigma$ is a symmetric positive semi-definite matrix. It is easy to check that the marginal distribution of the multivariate Laplace distribution for each dimension is the univariate Laplace distribution using the characteristic function in Eq.~\eqref{d-laplace}. The mean and variance of each marginal univariate Laplace distribution is the corresponding coordinate in the mean vector $m$ and the diagonal of $\Sigma$. If $m = 0$, the distribution has the following p.d.f. (see e.g. \cite{Kotz01, Wang08, Eltoft06}):
\begin{equation}
    \pi_x(x_1,...,x_d) = \frac{2}{(2\pi)^{d/2}|\Sigma|^{0.5}}\left(\frac{x^{\top}\Sigma^{-1}x}{2}\right)^{v/2}K_v\left(\sqrt{2x^{\top}\Sigma^{-1}x}\right),
\end{equation}
where $d$ is the number of dimensions, $v = (2-d)/2$, and $K_v$ is the modified Bessel function of the second kind. 
\begin{figure}
    \begin{subfigure}{.5\textwidth}
        \centering
        \includegraphics[width=1\linewidth, height=0.7\linewidth]{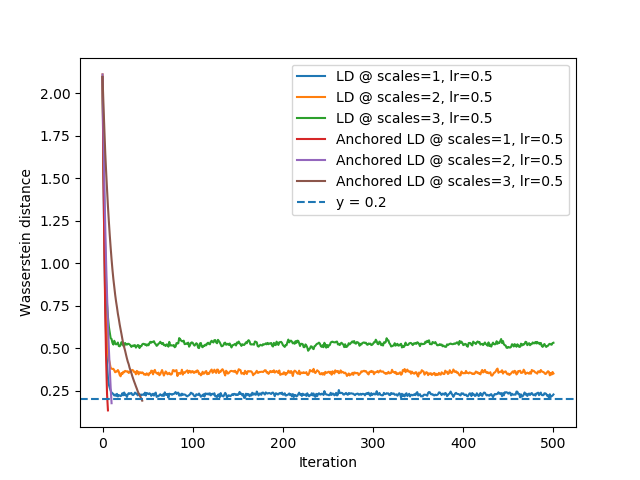}
        \caption{$\rho = 0$}
    \end{subfigure}
    \begin{subfigure}{.5\textwidth}
        \centering
        \includegraphics[width=1\linewidth, height=0.7\linewidth]{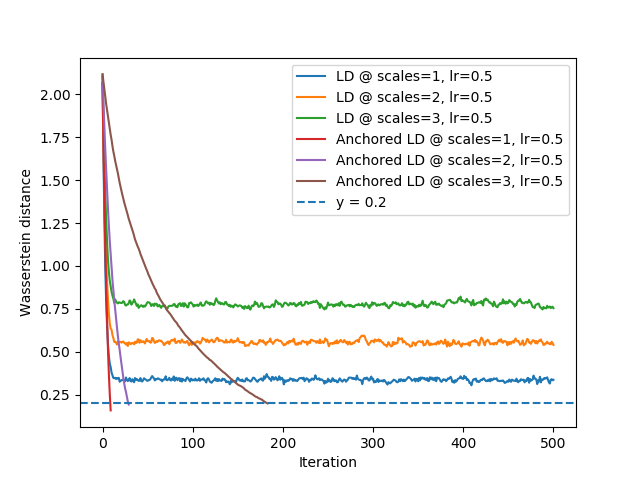}
        \caption{$\rho = 0.5$}
    \end{subfigure}
    \caption{Performance of the Langevin algorithms with Gaussian smoothing on simulating bivariate Laplace distribution from the prior distribution $\mathcal{N}(0,10I)$.}
    \label{2dLaplace}
\end{figure}
For the bivariate Laplace distribution, where $d = 2$ and the mean $m = 0$, less computational complexity is involved since the p.d.f. can be reduced to:
\begin{equation*}
    \pi_x(x_1,x_2) = \frac{1}{\pi \sigma_1 \sigma_2 \sqrt{1-\rho^2}}K_0\left( \sqrt{\frac{2}{1-\rho^{2}}\left(\frac{x_1^2}{\sigma_1^2}-\frac{2\rho x_1x_2}{\sigma_1\sigma_2}+\frac{x_2^2}{\sigma_2}\right)} \right),
\end{equation*}
where $\rho\in[-1,1]$ is a correlation coefficient. For the $d$-dimensional metric, to the best of our knowledge, there is no closed-form formula for the 2-Wasserstein distance between Laplace distributions. Hence, we use the sliced Wasserstein distance (see e.g. \cite{Nadjahi21}) as an estimate for our metric:
$SW^2_{2,L}(\nu_1,\nu_2) = \frac{1}{L}\sum^L_{l=1}\mathcal{W}^2_2\left(\nu_1^l,\nu_2^l\right)$,
where distributions $\nu_1$ and $\nu_2$ are projected onto $\mathbb{R}^d$ along $L$ directions. In our experiment, we will simply choose $L = d$ so that the squared sliced Wasserstein distance is the average of the squared distances of each dimension.

\begin{figure}
    \begin{subfigure}{.5\textwidth}
        \centering
        \includegraphics[width=1\linewidth, height=0.7\linewidth]{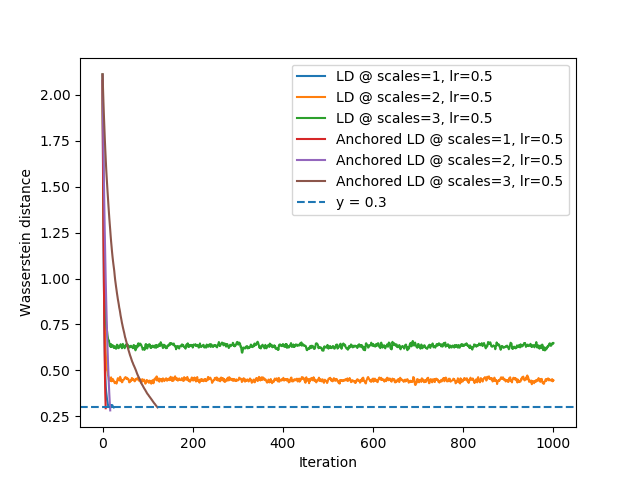}
        \caption{$\Sigma = I_d$}
    \end{subfigure}
    \begin{subfigure}{.5\textwidth}
        \centering
        \includegraphics[width=1\linewidth, height=0.7\linewidth]{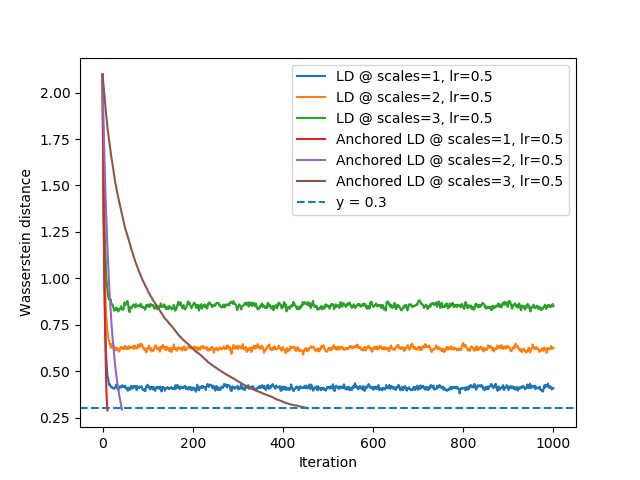}
        \caption{$\Sigma = I_d$ plus $\Sigma_{1,2} = \Sigma_{2,1} = 0.5$}
    \end{subfigure}
    \caption{Performance of the Langevin algorithms with Gaussian smoothing on simulating 3-dimensional Laplace distribution from the prior distribution $\mathcal{N}(0,10I)$.}
    \label{3dLaplace}
\end{figure}

We use the same hyper-parameters as univariate experiments. The marginal univariate Laplace distributions will have the standard deviations of $\sigma_1 = \sigma_2 = \cdots = \sigma_d = 1$. From the prior distribution $\mathcal{N}(0,10I)$, we report the results of the algorithms on simulating bivariate Laplace distributions in Figure~\ref{2dLaplace} and 3-dimensional Laplace distributions in Figure~\ref{3dLaplace}, both of which show that anchored Langevin algorithm achieves lower Wasserstein distance while the vanilla overdamped Langevin algorithm with Gaussian smoothing stops improving after some iterations. We summarize the results of Laplace simulation in Table \ref{Laplace:table}, which demonstrates the number of iterations needed by each model to reach Wasserstein distance less than a target $\epsilon$. The anchored Langevin algorithm shows superior performance especially at higher noise levels and higher stepsizes.
\begin{table}[h!]
    \begin{center}
        \caption{Number of iterations needed for Langevin algorithms to obtain 2-Wasserstein distance $<\epsilon$ while sampling Laplace distributions from the initial distribution $\mathcal{N}(0,10I)$. The results are averaged over 10 tries.}
        \label{Laplace:table}
        \begin{tabular}{ l | c c | c c | c c}
            Scale of random noise $\mu$ & \multicolumn{2}{c|}{1} & \multicolumn{2}{c|}{2} & \multicolumn{2}{c}{3} \\\hline
            Stepsize/ learning rate $\eta$ & 0.1 & 0.5 & 0.1 & 0.5 & 0.1 & 0.5 \\\hline
            \textbf{Univariate Laplace}, \boldmath$d = 1, \epsilon = 0.1$ & & & & & & \\
            \hspace{1cm} LD & 18.6 & $\infty^*$ & $\infty$ & $\infty$ & $\infty$ & $\infty$ \\
            \hspace{1cm} Anchored LD & 214.3 & 4.0 & 26.1 & 5.0 & 70.0 & 13.0 \\\hline
            \textbf{Bivariate Laplace}, \boldmath$d = 2, \epsilon = 0.2$ & & & & & & \\
            \hspace{1cm} LD, $\rho = 0$ & 25.3 & $\infty$ & 906.5 & $\infty$ & $\infty$ & $\infty$ \\
            \hspace{1cm} LD, $\rho = 0.5$ & 48.8 & $\infty$ & $\infty$ & $\infty$ & $\infty$ & $\infty$ \\
            \hspace{1cm} Anchored LD, $\rho = 0$ & 319.7 & 6.0 & 50.6 & 10.0 & 225.7 & 43.5 \\
            \hspace{1cm} Anchored LD, $\rho = 0.5$ & 43.5 & 9.0 & 147.2 & 29.3 & 929.7 & 183.2 \\\hline
            \textbf{Multivariate Laplace}, \boldmath$d = 3, \epsilon = 0.3$ & & & & & & \\
            \hspace{1cm} LD, $\Sigma = I_d$ & 30.5 & 22.1 & 86.0 & $\infty$ & $\infty$ & $\infty$ \\
            \hspace{1cm} LD, $\Sigma = I_d$ plus $\rho_{1,2} = 0.5$ & 45.4 & $\infty$ & $\infty$ & $\infty$ & $\infty$ & $\infty$ \\
            \hspace{1cm} Anchored LD, $\Sigma = I_d$ & 282.6 & 7.2 & 85.1 & 17.0 & 618.4 & 120.0 \\
            \hspace{1cm} Anchored LD, $\Sigma = I_d$ plus $\rho_{1,2} = 0.5$ & 49.4 & 11.0 & 220.0 & 42.8 & 2345.0 & 446.4 \\
            \multicolumn{7}{l}{$*$\footnotesize{ Wasserstein distance shows no sign of improving after a significant number of iterations}}
        \end{tabular}
    \end{center}
\end{table}

\subsection{Bayesian logistic regression on real data sets}
In this section, we will conduct Bayesian logistic regression on the Breast Cancer Wisconsin (Diagnostic) data set in the UCI Machine Learning Repository (\cite{Dua19}). In this data set, $X$ has $d=31$ dimensions and the data set contains $n=569$ samples, each of which describes the characteristics of the cell nuclei in a digitized image of a fine needle aspirate of a breast mass. The data is categorized into two classes with labels $y$. For the logistic regression, we will use the loss function with no bias:
\begin{equation}
    f(x) = -\frac{1}{n}\sum_{i=1}^{n}y_i\log\left(\sigma\left(x^{\top}X_i\right)\right) +(1-y_i)\log\left(1-\sigma\left(x^{\top}X_i\right)\right),
\end{equation}
where $\sigma(x)$ is the sigmoid function. Since there is no suitable Wasserstein distance metric for this experiment, we will use the prediction's accuracy from the algorithms instead. The accuracy will be averaged among $100$ independent runs. The initial distribution of the weights $x$ follows Laplace($0,b$) for $b > 0$, where we choose $b=2$. To add some non-differentiability to the loss function, we try some popular regularizers. We consider three different regularizers introduced as follows:
\begin{itemize}
    \item Lasso regularizer:
    $g(x) = \lambda\sum_{i=1}^{d}|x_i|$.
    \item Smoothly clipped absolute deviation (SCAD) regularizer (see e.g. \cite{Fan01}) is $g(x) = \sum_{i=1}^{d}p_{\lambda}(x_i)$ where $p_{\lambda}(x)$ is defined as (with $a>1$):
    \begin{equation}
        p_{\lambda}(x) = \begin{cases}
            \lambda |x| &\quad \text{if } |x| \leq \lambda, \\
            \frac{2a\lambda |x|-x^2-\lambda^2}{2(a-1)} &\quad \text{if } \lambda < |x| \leq a\lambda, \\
            \frac{\lambda^2(a+1)}{2} &\quad \text{otherwise}.
        \end{cases}
    \end{equation}
    \item Minimax concave penalty (MCP) regularizer (see e.g. \cite{Zhang10}) is $g(x) = \sum_{i=1}^{d}p_{\lambda}(x_i)$ where $p_{\lambda}(x)$ has the form:
    $p_{\lambda}(x) = 
            \lambda |x|-\frac{x^2}{2a}$ if $|x| \leq a\lambda$, and
            $p_{\lambda}(x) =\frac{a\lambda^2}{2}$ otherwise,
where $\lambda$ is the friction coefficient, and $a$ is the scaling coefficient. 
\end{itemize}

In Section~\ref{sec:analysis}, we introduced a special case of Bayesian logistic regression with mixed $\ell_2$-$\ell_1$ penalty of the form
$U(x) = f(x) + m_0\|x\|^2 + g(x)$, where $g(x)$ is the Lasso regularizer above.	For this regularizer, we can directly work with $U_{0}(x)=f(x)+ m_0\|x\|^2 + g^{\varepsilon}(x)$, 
where $g^{\varepsilon}$ is the smoothing of $\lambda\sum_{i=1}^{d}|x_{i}|$ defined as
$g^{\varepsilon}(x):=\lambda\sum_{i=1}^{d}\sqrt{x_{i}^{2}+\varepsilon^{2}}$ for some sufficiently small $\varepsilon$. The gradient of $U_0(x)$ is shown in Eq. \eqref{grad:special:case}. We can also replace $g(x)$ by the SCAD or MCP regularizer, whose smoothing versions are shown below with similar gradients.

For the SCAD regularizer, we can use the smoothed regularizer
$g^{\varepsilon}(x):=\sum_{i=1}^{d}p_{\lambda}^{\varepsilon}(x_{i})$, 
where
\begin{equation}\label{SCAD:smoothing}
    p_{\lambda}^{\varepsilon}(x)
    =\begin{cases}
        \lambda\sqrt{x^{2}+\varepsilon^{2}} &\text{if $|x|\leq \lambda$},
        \\ \frac{2\lambda\sqrt{a^{2}\lambda^{2}+\varepsilon^{2}}\sqrt{x^{2}+\varepsilon^{2}}-\lambda x^{2}-\lambda(\lambda^{2}+2\varepsilon^{2})}{2(\sqrt{a^{2}\lambda^{2}+\varepsilon^{2}}-\sqrt{\lambda^{2}+\varepsilon^{2}})} &\text{if $\lambda<|x|\leq a\lambda$},
        \\ 
        \frac{\lambda^{3}(a^{2}-1)}{2(\sqrt{a^{2}\lambda^{2}+\varepsilon^{2}}-\sqrt{\lambda^{2}+\varepsilon^{2}})} &\text{otherwise},
    \end{cases}
\end{equation}
where $a>1$. 
We can easily check that $p_{\lambda}^{\varepsilon}(x)$ in \eqref{SCAD:smoothing} is continuously differentiable.

For the MCP regularizer, we can use the smoothed regularizer
$g^{\varepsilon}(x):=\sum_{i=1}^{d}p_{\lambda}^{\varepsilon}(x_{i})$, 
where
\begin{equation}\label{MCP:smoothing}
    p_{\lambda}^{\varepsilon}(x)
    =\begin{cases}
        \lambda\sqrt{x^{2}+\varepsilon^{2}}-\frac{\lambda x^{2}}{2\sqrt{a^{2}\lambda^{2}+\varepsilon^{2}}} &\text{if $|x|\leq a\lambda$},
        \\ 
        \frac{\lambda(a^{2}\lambda^{2}+2\varepsilon^{2})}{2\sqrt{a^{2}\lambda^{2}+\varepsilon^{2}}} &\text{otherwise}.
    \end{cases}
\end{equation}
We can easily check that $p_{\lambda}^{\varepsilon}(x)$ in \eqref{MCP:smoothing} is continuously differentiable.
\begin{figure}
    \begin{subfigure}{.3\textwidth}
        \centering
        \includegraphics[width=1\linewidth]{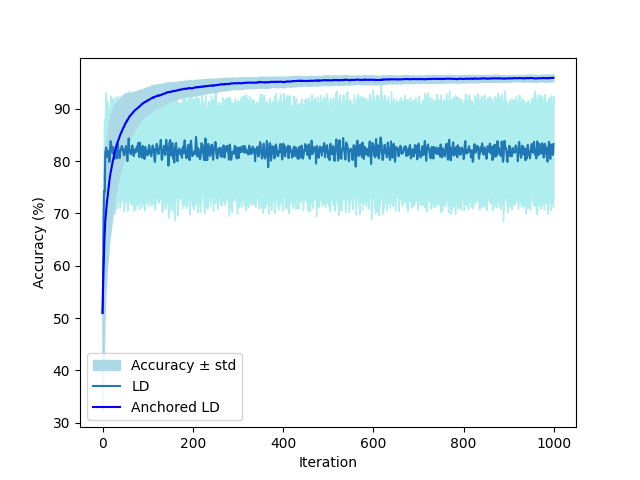}
        \caption{$\ell_2$-$\ell_1$}
    \end{subfigure}
    \begin{subfigure}{.3\textwidth}
        \centering
        \includegraphics[width=1\linewidth]{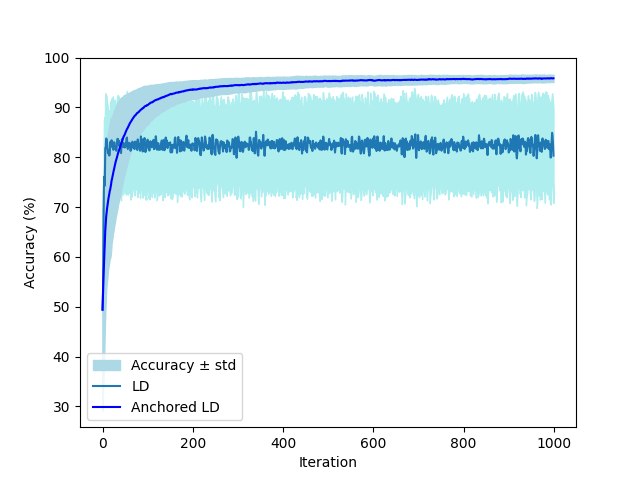}
        \caption{$\ell_2$-SCAD}
    \end{subfigure}
    \begin{subfigure}{.3\textwidth}
        \centering
        \includegraphics[width=1\linewidth]{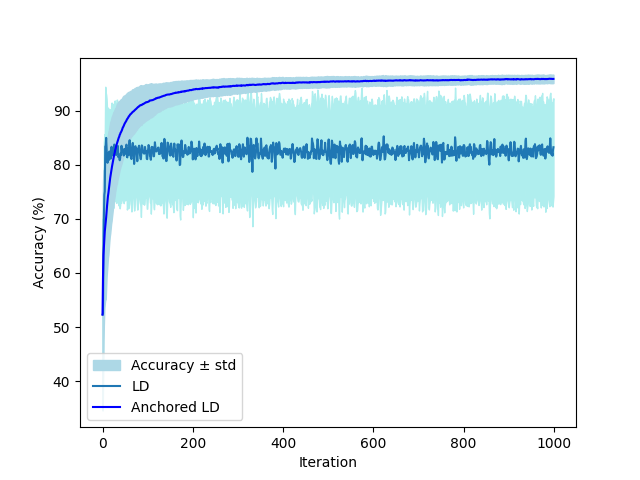}
        \caption{$\ell_2$-MCP}
    \end{subfigure}
    \caption{Performance of Bayesian logistic regression with mixed regularizers on the Breast Cancer Wisconsin data set using Langevin algorithms and deterministic smoothing. The accuracy is averaged over $100$ runs.}
    \label{Cancer:deterministic}
\end{figure}

For this experiment setup, we choose $m_0=0.5,\lambda=1,a=10$ and $\varepsilon=0.5$. Figure~\ref{Cancer:deterministic} shows that our algorithm outperforms and is more robust than the Langevin dynamics with $U_0(x)$ replacing $U(x)$ for all types of regularizers.

\begin{figure}
    \begin{subfigure}{.3\textwidth}
        \centering
        \includegraphics[width=1\linewidth]{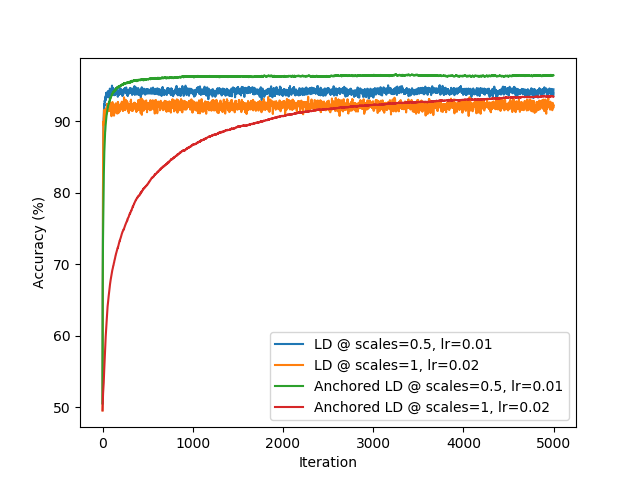}
        \caption{Lasso}
    \end{subfigure}
    \begin{subfigure}{.3\textwidth}
        \centering
        \includegraphics[width=1\linewidth]{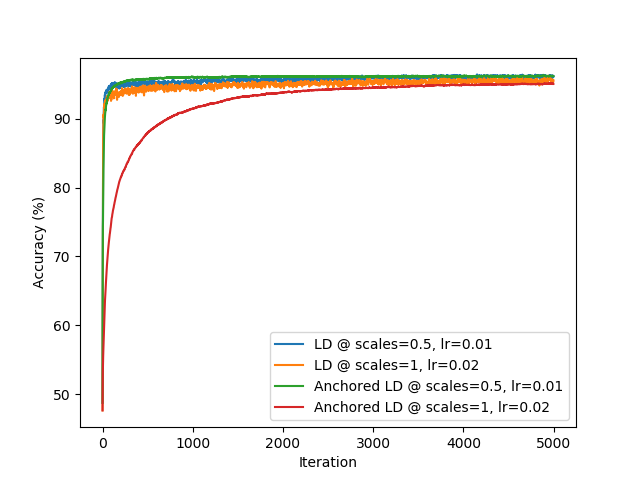}
        \caption{SCAD}
    \end{subfigure}
    \begin{subfigure}{.3\textwidth}
        \centering
        \includegraphics[width=1\linewidth]{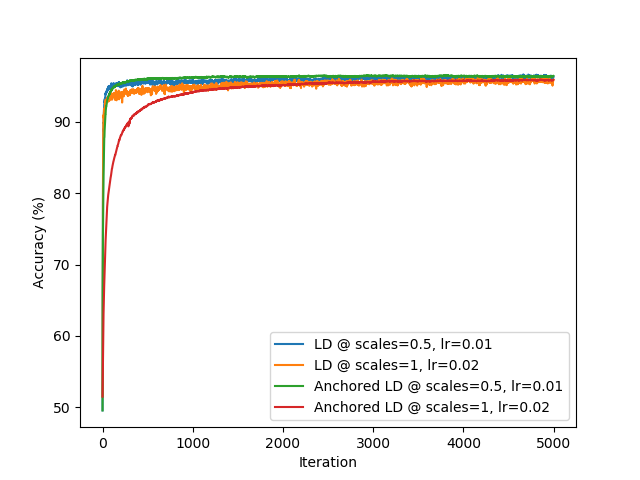}
        \caption{MCP}
    \end{subfigure}
    \caption{Performance of Bayesian logistic regression on the Breast Cancer Wisconsin data set using Langevin algorithms with Gaussian smoothing. The accuracy is averaged over $100$ runs.}
    \label{Cancer:Gauss}
\end{figure}

For the Gaussian smoothing experiments, we will use the following loss function $U(x)$:
\begin{align*}
    U(x) = f(x) + g(x) = -\frac{1}{n}\sum_{i=1}^{n}y_i\log\left(\sigma\left(x^{\top}X_i\right)\right) +(1-y_i)\log\left(1-\sigma\left(x^{\top}X_i\right)\right)+g(x),
\end{align*}
where $g(x)$ is the regularizer. With the same setup as in Section~\ref{section:smooth}, $f(x)$ is the smooth component of the loss function, and $g(x)$ is a non-smooth function with the Gaussian smoothing $g_0(x)$ as the reference potential. In this experiment, we choose the same hyper-parameters as the deterministic smoothing experiments: $a = 10$, $\lambda=1$. All expectations are approximated by $500$ Monte Carlo simulations each. Figure~\ref{Cancer:Gauss} shows the result of Bayesian logistic regression with Lasso, SCAD and MCP regularizations on the Breast Cancer Wisconsin data set.

\subsection{Feedforward neural network}

\begin{figure}
    \begin{subfigure}{.45\textwidth}
        \centering
        \includegraphics[width=1\linewidth, height=0.7\linewidth]{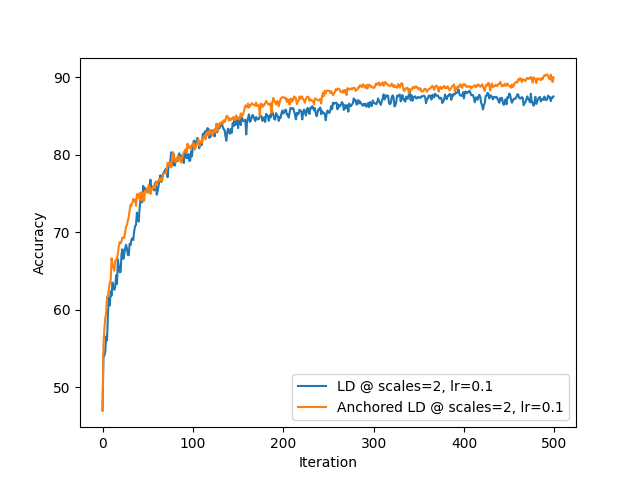}
        \caption{Accuracy}
    \end{subfigure}
    \begin{subfigure}{.45\textwidth}
        \centering
        \includegraphics[width=1\linewidth, height=0.7\linewidth]{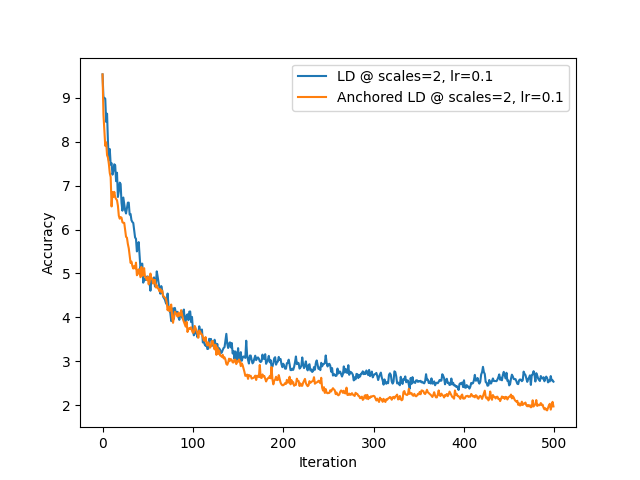}
        \caption{Loss}
    \end{subfigure}
    \caption{Performance of two-layer feedforward neural network on the Breast Cancer Wisconsin data set. The accuracy and loss value are averaged over $50$ runs.}
    \label{NN1}
\end{figure}

\begin{figure}
    \begin{subfigure}{.45\textwidth}
        \centering
        \includegraphics[width=1\linewidth, height=0.7\linewidth]{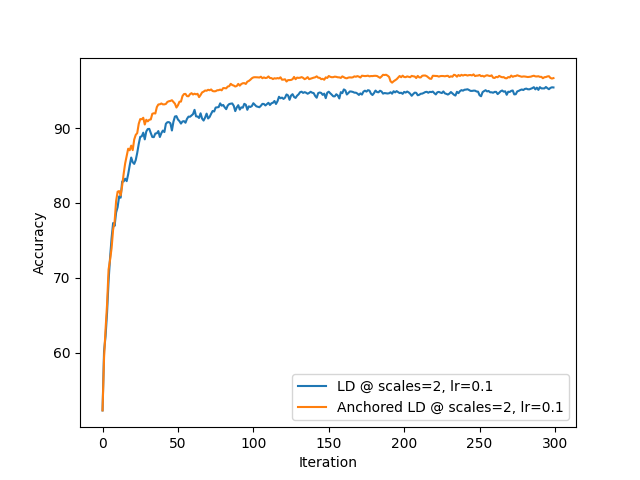}
        \caption{Accuracy}
    \end{subfigure}
    \begin{subfigure}{.45\textwidth}
        \centering
        \includegraphics[width=1\linewidth, height=0.7\linewidth]{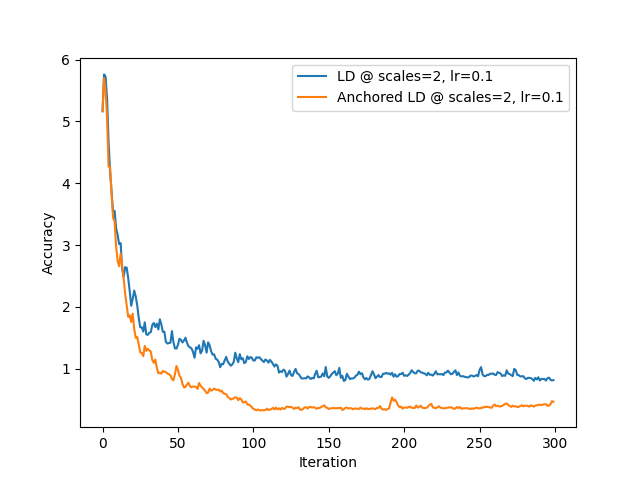}
        \caption{Loss}
    \end{subfigure}
    \caption{Performance of two-layer feedforward neural network on the Banknote Authentication data set. The accuracy and loss value are averaged over $50$ runs.}
    \label{NN2}
\end{figure}

In this section, we test the algorithms on a neural network with two layers, where the first layer uses ReLU activation function and the second layer uses sigmoid activation. We use binary cross entropy as the loss function. The second layer has one node to output the predicted probability and let $n=32$ be the number of nodes in the first layer. The loss function of this neural network is:
\begin{align}
    U\left(w_1,w_2\right) & = -\frac{1}{N}\sum\nolimits_{i=1}^{N}y_i\log\left(g\left(w_1,w_2\right)\right) + (1-y_i)\log\left(1-g\left(w_1,w_2\right)\right), \\
    g\left(w_1,w_2\right) & = \sigma\left(\sum\nolimits_{j=1}^{n}f\left(X.w_{1_j}\right).w_{2_j}\right),
\end{align}
where $w_1, w_2$ are the weights in the first and second layers, $f(x)$ is the ReLU activation function, $\sigma(x)$ is the sigmoid function and $g(w_1,w_2)$ is the output of the neural network. Since our emphasis is to solve the non-differentiability problem of ReLU activation function, the gradients of ReLU layer's weights are approximated with Gaussian smoothing and updated by Langevin algorithms. The sigmoid function is differentiable, thus the second layer's weights are updated by the conventional overdamped Langevin dynamics due to the gradients accessibility. Each expectation will be estimated by 200 Monte Carlo simulations. Let the prior distributions of $w_1$ and $w_2$ be $\mathcal{N}(0,4)$. We run the experiment on the Breast Cancer Wisconsin data set and also the Banknote Authentication data set in the UCI Machine Learning Repository (\cite{Dua19}). The Banknote Authentication data set has $n=1372$ samples with dimension $d=4$, which were extracted from images taken from genuine and forged banknote-like specimens. Figures~\ref{NN1} and \ref{NN2} show the accuracy and loss value of the above neural network on the two data sets using Langevin algorithms, where our anchored LD achieves better performance.



\subsection{Sampling heavy-tailed distributions}

In this section, we will demonstrate that anchored Langevin algorithms outperform the overdamped Langevin algorithm in sampling heavy-tailed distributions. In Theorem~\ref{thm1} and Example~\ref{ex:student}, we refer to the following function as an example of heavy-tailed distributions. 
Consider the Gibbs distribution
$\pi(x)\propto e^{-U(x)}$,
with potential
\begin{equation}\label{formula:heavytail}
U(x)=\iota\log\!\bigl(1+\Vert x \Vert^{2}\bigr),
\qquad
\iota>1+\frac{d}{2}.
\end{equation}
By construction
$e^{-U(x)}=(1+\Vert x \Vert^{2})^{-\iota}$,
which satisfies the heavy tail behavior
as $\Vert x \Vert\to\infty$.

\begin{figure}
    \begin{subfigure}{.5\textwidth}
        \centering
        \includegraphics[width=0.9\linewidth, height=0.6\linewidth]{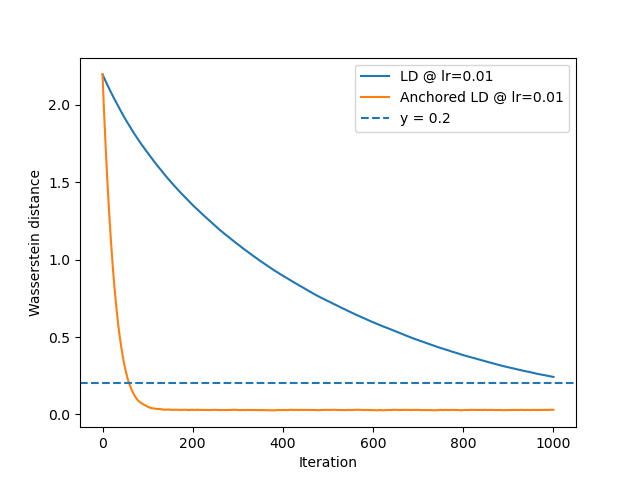}
        \caption{Prior $\mathcal{N}(0,10)$}
        \label{heavytaileda}
    \end{subfigure}
    \begin{subfigure}{.5\textwidth}
        \centering
        \includegraphics[width=0.9\linewidth, height=0.6\linewidth]{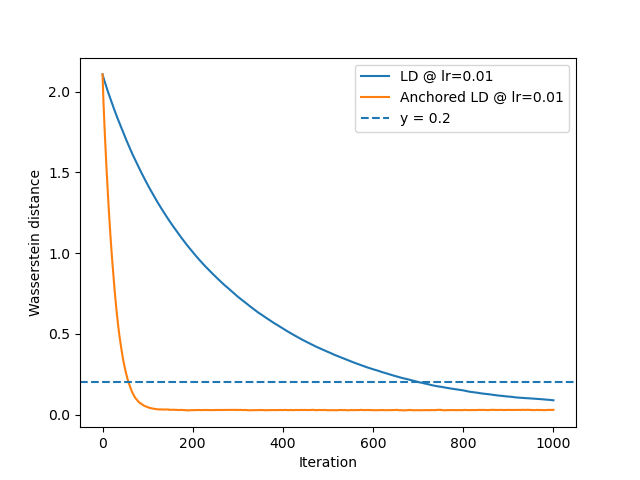}
        \caption{Prior Uniform$(-5,5)$}
        \label{heavytailedb}
    \end{subfigure}
    \caption{Performance of the anchored Langevin algorithm on sampling heavy-tailed distribution $\pi(x)\;\propto\;e^{-U(x)}$.}
    \label{heavytailed}
\end{figure}

Since the potential~\eqref{formula:heavytail} is differentiable, we choose a suitable reference potential instead of the Gaussian smoothing method to avoid the high cost of Monte Carlo simulations in approximating expectations. For any $\beta>\tfrac{d}{2}$ define the reference potential
$U_{0}(x):= \beta\log(1+\lVert x\rVert^{2})$.
Similar to the setup of the Laplace distribution simulation, we sample $5,000$ data points and estimate the 2-Wasserstein distance between the simulated sample and the true distribution. 2-Wasserstein distance can be estimated using the quantile function of the heavy-tailed distributions. We choose the following hyper-parameters for $U(x)$ and $U_{0}(x)$: $\iota=2$, $\beta=1 < \iota$, and stepsize $\eta=0.01$. We test two different prior distributions, $\mathcal{N}(0,10I)$ and Uniform$(-5,5)$. To reduce variations, we average 2-Wasserstein distance over $100$ repeated runs. Figure~\ref{heavytailed} shows that our anchored Langevin algorithm converges much faster compared to the overdamped Langevin dynamics if we choose a suitable reference potential function $U_{0}(x)$ for the target heavy-tailed distribution.

\section{Conclusion}

First-order Langevin algorithms such as ULA have become standard tools for large-scale sampling, yet their reliance on differentiable log-densities and their poor performance on heavy-tailed targets limit their applicability. We introduced anchored Langevin dynamics, a general framework that addresses both issues by incorporating a smooth reference potential and modifying the Langevin diffusion through multiplicative scaling. Our theoretical analysis established non-asymptotic convergence guarantees in the 2-Wasserstein distance and revealed an equivalent random time-change formulation. Empirical results further demonstrated that anchored Langevin dynamics can effectively handle non-smooth and heavy-tailed targets. Taken together, these contributions highlight anchored Langevin dynamics as a principled and practical alternative to existing first-order methods, and open up new directions for scalable sampling algorithms in challenging settings.

\acks{The authors would like to thank Qi Feng, Jin Ma, Molei Tao, Jianfeng Zhang for helpful discussions. 
Mert G\"urb\"uzbalaban's research is supported in part by
the grants Office of Naval Research Award Numbers N00014-21-1-2244 and N00014-24-1-2628, National Science Foundation (NSF) CCF-1814888, NSF DMS-2053485. Hoang M. Nguyen and Lingjiong Zhu are partially supported by the grants NSF DMS-2053454, NSF DMS-2208303. Xicheng Zhang is partially supported by National Key R\&D program of China (No. 2023YFA1010103) and NNSFC grant of China (No. 12131019).
}




\appendix

\section{Technical Background}

We present a review of some technical background in probability theory,
and the discussions about Markov semigroup, infinitesimal generator, reversibility and Wasserstein metric can be found in e.g. \cite{EthierKurtz}, \cite{RevuzYor} and \cite{villani2008optimal}. 

\begin{itemize}
\item 
\textbf{Markov semigroup.}
Given a Markov process $(X_{t})_{t\geq 0}$ on $\mathbb{R}^{d}$, 
the Markov semigroup $(P_{t})_{t\geq 0}$ is
defined on $C(\mathbb{R}^{d})$, the space of bounded continuous functions
on $\mathbb{R}^{d}$ via:
$P_{t}(f(x)):=\mathbb{E}[f(X_{t})|X_{0}=x]$.
By the Markov property, $P_{t+s}(f)=P_{t}(P_{s}(f))$ for any $t,s\geq 0$
and hence $(P_{t})_{t\geq 0}$ forms a semigroup.
\item
\textbf{Infinitesimal generator.}
The infinitesimal generator $\mathcal{L}$
of a Markov semigroup $(P_{t})_{t\geq 0}$ is defined by
$\mathcal{L}f:=\lim_{t\downarrow 0}\frac{P_{t}(f)-f}{t}$
for all $f\in\mathcal{D}(\mathcal{L})$, where $\mathcal{D}(\mathcal{L})$
is the subset of $C(\mathbb{R}^{d})$ where this limit exists. 
\item
\textbf{Reversibility.}
Let $P_{t}$ be the associated Markov semigroup of a Markov process $(X_{t})_{t\geq 0}$ on $\mathbb{R}^{d}$. 
A probability measure $\pi$ is reversible with respect to $(P_{t})_{t\geq 0}$ if
$\int_{\mathbb{R}^{d}}f\mathcal{L}gd\pi=\int_{\mathbb{R}^{d}}g\mathcal{L}fd\pi$,
for any $f,g\in\mathcal{D}(\mathcal{L})$. 
\item
\textbf{Wasserstein metric.}
For any $p\geq 1$, define $\mathcal{P}_{p}(\mathbb{R}^{d})$
as the space consisting of all the Borel probability measures $\nu$
on $\mathbb{R}^{d}$ with the finite $p$-th moment
(based on the Euclidean norm).
For any two Borel probability measures $\nu_{1},\nu_{2}\in\mathcal{P}_{p}(\mathbb{R}^{d})$, 
we define the standard $p$-Wasserstein
metric:
$\mathcal{W}_{p}(\nu_{1},\nu_{2}):=\left(\inf\mathbb{E}\left[\Vert Z_{1}-Z_{2}\Vert^{p}\right]\right)^{1/p},$
where the infimum is taken over all joint distributions of the random variables $Z_{1},Z_{2}$ with marginal distributions
$\nu_{1},\nu_{2}$.
\end{itemize}


\section{Technical Proofs}

\subsection*{Proof of Theorem~\ref{thm:continuous:time}}
	Notice that the infinitesimal generator of the anchored Langevin SDE \eqref{SDE0} is given by
	$\mathcal{L}:=\sigma^2\Delta+b\cdot\nabla$.
	By \eqref{CN1}, we have
	\begin{align*}
		\mathcal{L} \Vert x\Vert^2=2\<x,b(x)\>+2d\sigma^2(x)&=2\Big[d-\<x, \nabla U_0(x)\>\Big] e^{(U-U_0)(x)}\leq -2c_0\Vert x\Vert^{2+r}+2c_1.
	\end{align*}
	Since $\sigma$ is positive and continuous, it is well-known that there is a unique weak solution
	to SDE \eqref{SDE0} (see e.g. \cite{SV79}).
	Let $\mathcal{L}^*$ be the adjoint operator of $\mathcal{L}$.
	By definition \eqref{11}, we have
	$$
	\mathcal{L}^* e^{-U}=\Delta\left(\sigma^2 e^{- U}\right)-\div\left(b e^{-U}\right)
	=\Delta e^{-U_0}+\div\left(\nabla U_0 e^{-U_0}\right)\equiv0.
	$$
	Hence, 
	\begin{align*}
		\partial_t\int_{\mathbb{R}^d} P_tf(x)\pi(dx)=\frac1M\int_{\mathbb{R}^d}\mathcal{L} P_tf(x) e^{-U(x)}dx=
		\frac1M\int_{\mathbb{R}^d}P_tf(x)\mathcal{L}^* e^{-U(x)}dx=0,
	\end{align*}
	where $M:=\int_{\mathbb{R}^{d}}e^{-U(x)}dx$ and $P_{t}$ is the semigroup defined by the anchored Langevin SDE \eqref{SDE0}.
	From this we deduce that
	$\int_{\mathbb{R}^d} P_tf(x)\pi(dx)=\int_{\mathbb{R}^d}f(x)\pi(dx)$.
	That is, $\pi$ is an invariant measure of $P_t$. 
	Moreover, by Theorem 7.4 in \cite{XZ20}, $\pi$ is the unique invariant measure and (i) and (ii) hold. This completes the proof.
\hfill $\Box$


\subsection*{Proof of Theorem~\ref{thm1}}

With the choice $U_0(x):=\beta\log q(x)$
and $U(x):=(\beta+1)\log q(x)$, one can compute
$$
[d-\<x, \nabla U_0(x)\>] e^{(U-U_0)(x)}
=dq(x)-\beta\<x,\nabla q(x)\>.
$$
Thus, Assumption~\ref{assump:continuous} is satisfied.
By Theorem~\ref{thm:continuous:time}, $q(x)^{-1-\beta}/\int_{\R^d}q(x)^{-1-\beta}dx$ is the unique stationary distribution of the SDE \eqref{overdamped:SDE:modified}.
This completes the proof.
\hfill $\Box$

%


\subsection*{Proof of Lemma~\ref{prop:reversible}}
	The infinitesimal generator of the anchored Langevin SDE \eqref{overdamped:SDE:modified}
	is given by
	\begin{equation}
		\mathcal{L}=e^{U(x)-U_{0}(x)}\Delta-e^{U(x)-U_{0}(x)}\nabla U_{0}(x)\cdot\nabla
		=e^{U(x)-U_{0}(x)}\mathcal{L}_{0},
	\end{equation}
	where $\mathcal{L}_{0}$ is the infinitesimal generator of
	the overdamped Langevin SDE:
	\begin{equation}
		dX_{t}=-\nabla U_{0}(X_{t})dt+\sqrt{2}dW_{t},
	\end{equation}
	which admits a unique invariant distribution
	$\pi_{0}\propto e^{-U_{0}(x)}$, 
	whereas \eqref{overdamped:SDE:modified}
	admits a unique invariant distribution
	$\pi\propto e^{-U(x)}$
	so that $e^{U-U_{0}}d\pi=\frac{\int_{\mathbb{R}^{d}}e^{-U_{0}(x)}dx}{\int_{\mathbb{R}^{d}}e^{-U(x)}dx}d\pi_{0}$.
	
	For any $f,g\in\mathcal{D}(\mathcal{L})$, we can compute that
	\begin{equation}\label{together:1}
		\int_{\mathbb{R}^{d}}f\mathcal{L}(g)d\pi
		=\int_{\mathbb{R}^{d}}fe^{U-U_{0}}\mathcal{L}_{0}(g)d\pi
		=\frac{\int_{\mathbb{R}^{d}}e^{-U_{0}(x)}dx}{\int_{\mathbb{R}^{d}}e^{-U(x)}dx}\int_{\mathbb{R}^{d}}f\mathcal{L}_{0}(g)d\pi_{0},
	\end{equation}
	and it is well known that the overdamped Langevin SDE is reversible (see e.g. \cite{ChenICLR2019})
	so that
	\begin{equation}\label{together:2}
		\int_{\mathbb{R}^{d}}f\mathcal{L}_{0}(g)d\pi_{0}
		=\int_{\mathbb{R}^{d}}\mathcal{L}_{0}(f)gd\pi_{0},
	\end{equation}
	and moreover,
	\begin{equation}\label{together:3}
		\int_{\mathbb{R}^{d}}\mathcal{L}_{0}(f)gd\pi_{0}
		=\int_{\mathbb{R}^{d}}e^{U_{0}-U}\mathcal{L}(f)gd\pi_{0}
		=\frac{\int_{\mathbb{R}^{d}}e^{-U(x)}dx}{\int_{\mathbb{R}^{d}}e^{-U_{0}(x)}dx}\int_{\mathbb{R}^{d}}\mathcal{L}(f)gd\pi,
	\end{equation}
	and together from \eqref{together:1}, \eqref{together:2}, \eqref{together:3}
	we conclude that
	$\int_{\mathbb{R}^{d}}f\mathcal{L}(g)d\pi=\int_{\mathbb{R}^{d}}\mathcal{L}(f)gd\pi$,
	and hence the anchored Langevin SDE \eqref{overdamped:SDE:modified} is reversible.
	This completes the proof.
\hfill $\Box$


\subsection*{Proof of Lemma~\ref{lem:Dirichlet}}
We can compute that
\begin{equation}
		\mathcal{E}(f)=\frac{1}{2}\int_{\mathbb{R}^{d}}\left(\mathcal{L}(f^{2})-2f\mathcal{L}(f)\right)d\pi
		=\frac{1}{2}\int_{\mathbb{R}^{d}}e^{U-U_{0}}\left(\mathcal{L}_{0}(f^{2})-2f\mathcal{L}_{0}(f)\right)d\pi,
\end{equation}
and moreover, it is easy to compute that
\begin{equation}
		\mathcal{L}_{0}(f^{2})-2f\mathcal{L}_{0}(f)
		=\Delta f^{2}-\nabla U_{0}\cdot\nabla f^{2}
		-2f\left(\Delta f-\nabla U_{0}\cdot\nabla f\right)
		=2\Vert\nabla f\Vert^{2},
\end{equation}
which yields the desired result and completes the proof.
\hfill $\Box$


\subsection*{Proof of Proposition~\ref{prop:chi:square}}
Since we proved in Lemma~\ref{prop:reversible} that $(X_{t})_{t\geq 0}$ is reversible, 
we have
$\frac{d}{dt}\chi^{2}(\mu_{t}\Vert\pi)=-2\mathcal{E}\left(\frac{d\mu_{t}}{d\pi}\right)$,
where
$\mathcal{E}(f):=-\int_{\mathbb{R}^{d}}f\mathcal{L}(f)d\pi$
is the Dirichlet form. 
By Lemma~\ref{lem:Dirichlet}, we have
\begin{equation}
	\frac{d}{dt}\chi^{2}(\mu_{t}\Vert\pi)=-2\int_{\mathbb{R}^{d}}e^{U-U_{0}}\left\Vert\nabla\left(\frac{d\mu_{t}}{d\pi}\right)\right\Vert^{2}d\pi.
\end{equation}
On the other hand, let $\tilde{\mu}_{t}$ denote the distribution of the overdamped Langevin SDE $(X_{t})_{t\geq 0}$ 
in \eqref{overdamped:SDE}.
Then, we have
\begin{equation}
	\frac{d}{dt}\chi^{2}(\tilde{\mu}_{t}\Vert\pi)=-2\int_{\mathbb{R}^{d}}\left\Vert\nabla\left(\frac{d\tilde{\mu}_{t}}{d\pi}\right)\right\Vert^{2}d\pi.
\end{equation}
If $\pi$ satisfies a Poincar\'{e} inequality with constant $C_{P}$; see e.g. \cite{Bakry2008,Bakry2013}, then for any $\nu\ll\pi$, 
\begin{equation}
	\chi^{2}(\nu\Vert\pi)\leq C_{P}\cdot\mathcal{E}\left(\frac{d\nu}{d\pi}\right).
\end{equation}
It follows that for the overdamped Langevin SDE $(X_{t})_{t\geq 0}$ 
in \eqref{overdamped:SDE}, 
\begin{equation}
	\chi^{2}(\tilde{\mu}_{t}\Vert\pi)\leq
	\chi^{2}(\tilde{\mu}_{0}\Vert\pi)e^{-2t/C_{P}}.
\end{equation}
If $U(x)\geq U_{0}(x)$, then 
for the anchored Langevin SDE $(X_{t})_{t\geq 0}$ 
in \eqref{overdamped:SDE:modified},
\begin{equation}
	\frac{d}{dt}\chi^{2}(\mu_{t}\Vert\pi)
	\leq
	-2e^{\inf_{x\in\mathbb{R}^{d}}(U(x)-U_{0}(x))}
	\int_{\mathbb{R}^{d}}\left\Vert\nabla\left(\frac{d\mu_{t}}{d\pi}\right)\right\Vert^{2}d\pi,
\end{equation}
so that
$\chi^{2}(\mu_{t}\Vert\pi)
\leq\chi^{2}(\mu_{0}\Vert\pi)e^{-2at/C_{P}}$,
where
$a:=e^{\inf_{x\in\mathbb{R}^{d}}(U(x)-U_{0}(x))}$ provided that $a\in(0,\infty)$.
This completes the proof.
\hfill $\Box$


\subsection*{Proof of Lemma~\ref{lem:crucial}}

By \eqref{DIS}, it is well-known that $Z_t$ is Harris recurrent (see \cite{MT2009}), i.e.,
$$
\int^\infty_01_{\{\|Z_s\|\leq 1\}}ds=\infty,\ \ a.s.
$$
From this we derive that
$$
\int^\infty_0 e^{(U_0-U)(Z_s)}ds\geq \inf_{\|x\|\leq 1}e^{(U_0-U)(x)}\int^\infty_0 1_{\{\|Z_s\|\leq 1\}}ds=\infty.
$$
Hence, $\ell(t)$ is finite for all $t>0$ and
$\int^{\ell(t)}_0 e^{(U_0-U)(Z_s)}ds=t$.
By differentiating both hand sides with respect to $t$ and applying the chain rule, we get
$\ell'(t)e^{(U_0-U)(Z_{\ell(t)})}=1$.
The proof is complete.
\hfill $\Box$


\subsection*{Proof of Theorem~\ref{thm:time:change}}
	By the change of variable, we have
	\begin{align*}
		X_t=x-\int^{\ell(t)}_0\nabla U_0(Z_s)ds+\sqrt{2}\widetilde W_{\ell(t)}
          =x-\int^t_0\nabla U_0(X_s)\ell'(s)ds+\sqrt{2}\widetilde W_{\ell(t)}.
	\end{align*}
	If we define
	$$
W_t:=\int^t_0\sqrt{1/\ell'(s)}d\widetilde W_{\ell(s)},
	$$
	then the covariation between $W^{i}$ and $W^{j}$ can be computed as
	$$
	\left\langle W^i, W^j\right\rangle_t=\delta_{i=j}\int^t_01/\ell'(s)d\ell(s)=t\delta_{i=j},
	$$
	where $W=(W^{1},W^{2},\ldots,W^{d})$.
	Hence, by L\'evy's characterization of Brownian motion (see e.g. \cite{RevuzYor}), $W_t$ is still a Brownian motion, and
	$\widetilde W_{\ell(t)}=\int^t_0\sqrt{\ell'(s)}dW_s$.
By \eqref{ODE1}, we get	
\begin{align*}
		X_t&=x-\int^t_0\nabla U_0(X_s) e^{(U-U_0)(X_s)}ds+\sqrt{2}\int^t_0 e^{(U-U_0)(X_s)/2}dW_s.
	\end{align*}
	In particular, $X_t$ is a solution of SDE \eqref{SDE1}.  
	The uniqueness is well known (see e.g. \cite{SV79}). The proof is complete.
\hfill $\Box$




\subsection*{Proof of Corollary~\ref{cor:heavy:tailed:example:2}}

With the choice $U_0(x):=\beta\log q(x)$
and $U(x):=(\beta+1)\log q(x)$, one can compute
\begin{align*}
&b(x)=-\nabla U_0(x)e^{(U-U_0)(x)}=-\beta\nabla q(x),
\\
&\sigma(x)=e^{(U-U_0)(x)/2}=\sqrt{q(x)},
\end{align*}
and furthermore, one can check that Assumption~\ref{assump:b:sigma} is satisfied.

By Theorem~\ref{thm:continuous:time}, $q(x)^{-1-\beta}/c_3$ is the unique stationary distribution of the SDE \eqref{overdamped:SDE:modified}.
This completes the proof.
\hfill $\Box$


\subsection*{Proof of Theorem~\ref{thm:discrete}}
By applying the supporting lemmas in Appendix~\ref{sec:support:lemmas} and applying Theorem 3.3. and Theorem 3.4 from \cite{Tao2021}, we have that
	for any $0<\eta\leq\eta_{\max}$, 
	\begin{equation}
		\mathbb{E}\Vert X_{\eta k}-x_{k}\Vert^{2}\leq C^{2}\eta^{2p_{2}-1},
	\end{equation}
	and
	\begin{equation}
		\mathcal{W}_{2}(\nu_{k},\pi)\leq\sqrt{2}e^{-\beta k\eta}\mathcal{W}_{2}(\nu_{0},\pi)+\sqrt{2}C\eta^{p_{2}-\frac{1}{2}},
	\end{equation}
	where
	\begin{equation}
		\eta_{\max}:=\min\left\{t_{0},\eta_{1},\eta_{2},\frac{1}{4\beta},\left(\frac{\sqrt{\beta}}{4\sqrt{2}D_{2}}\right)^{\frac{1}{p_{2}-\frac{1}{2}}},
		\left(\frac{\beta}{8\sqrt{2}(D_{1}+C_{0}D_{2})}\right)^{\frac{1}{p_{2}-\frac{1}{2}}}\right\},
	\end{equation}
	and
	\begin{equation}
		C:=\frac{2}{\sqrt{\beta}}\left(\frac{C_{1}+C_{0}C_{2}+\sqrt{2}C_{3}(D_{1}+C_{0}D_{2})}{\sqrt{\beta}}+C_{2}+\sqrt{2}D_{2}C_{3}\right),
	\end{equation}
	where 
   $C_{3}:=\sqrt{4\mathbb{E}\Vert X_{0}\Vert^{2}+6\mathbb{E}_{X\sim\pi}\Vert X\Vert^{2}}$.
   
The proof is complete by recalling from the supporting lemmas in Appendix~\ref{sec:support:lemmas}
that $\beta=m-\alpha$, $C_{0}=4L$, $t_{0}=\frac{1}{L^{2}+4\alpha}$, $C_{1}=3L\sqrt{1+4\alpha}\left(\Vert x_{\ast}\Vert+\Vert\sigma(x_{\ast})I_{d}\Vert_{\mathrm{HS}}\right)$, $D_{1}=2L\sqrt{1+4\alpha}$, 
$C_{2}=7(1+4\alpha)\left(\Vert x_{\ast}\Vert+\Vert\sigma(x_{\ast})I_{d}\Vert_{\mathrm{HS}}\right)$,
$D_{2}:=5(1+4\alpha)$,
and we can take $p_{1}=3/2$, $p_{2}=1$, $\eta_{1}=\frac{1}{L^{2}+4\alpha}$, $\eta_{2}=\frac{1}{L^{2}+4\alpha}$.
\hfill $\Box$


\subsection*{Proof of Theorem~\ref{thm:equivalence}}
	Indeed, since $x_k = z_{\ell_k}$ for any $k$, Eq. \eqref{time-change} can be written as:
	\begin{align*}
		x_{k+1} & = x_k - \eta\exp\left\{U(x_k)-U_0(x_k)\right\}\nabla U_0(x_k)+ \left(2\eta\exp\left\{U(x_k)-U_0(x_k)\right\}\right)^{1/2}\xi_{k+1} \\
		& = x_k + \eta b(x_k)+\sqrt{2\eta}\sigma(x_k)\xi_{k+1},
	\end{align*}
	which is Eq. \eqref{discrete:dynamics}. Hence, with the same initial $x_0$ and synchronous noise $\xi_k$, the two discretizations are equivalent.
		This completes the proof.
\hfill $\Box$

\subsection*{Proof of Lemma~\ref{lem:CP:bound}}
    By the definition of $U_{0}$, we have
    $$
    |U(x)-U_0(x)|=|\mathbb{E}[g(x+\mu\xi)]-g(x)|\leq K\mathbb{E}\|\mu\xi\|
    \leq 
    K\mu\left(\mathbb{E}\|\xi\|^{2}\right)^{1/2}
    =K\mu\sqrt{d}.
    $$
    This completes the proof.
\hfill $\Box$

\subsection*{Proof of Lemma~\ref{ineq-small-grad-error}} 
First of all, note that by \eqref{eq-grad-smoothed} we have 
$\nabla g_0(x) = \frac{1}{\mu}\mathbb{E}_{\hat{\xi}}\left[g\left(x+\mu\hat{\xi}\right)\hat{\xi}\right]$,
where $\hat{\xi}\sim\mathcal{N}(0,I_{d})$. By our assumptions on $g$, for $x,y\in \mathbb{R}^d$ and $u\in\partial g(x)$ we can write
 \begin{equation}
 g(y) = g(x) + \langle u, y-x \rangle + o(x,y,u), 
 \label{eq-subgrad=estimate}
\end{equation}
with 
 \begin{equation}
      \sup_{u\in \partial g(y)} \big|{o(x,y,u)}\big| \leq C_{g} \|y-x\|^{2}.
       \label{subgrad-bound-3}
 \end{equation}
Given $u \in \partial g(x)$, applying \eqref{eq-subgrad=estimate} with $y=x+\mu\hat{\xi}$, we have
 \begin{align} 
 \nabla g_0(x) -  u = \mathbb{E}_{\hat{\xi}}\left[
 \frac{1}{\mu}\left(
 g(x) + \mu \left\langle u, \hat\xi \right\rangle + o\left(x, x+\mu\hat\xi, u\right) \right)
 \hat{\xi} - u
 \right]= \frac{\mathbb{E}_{\hat{\xi}} \left[ o\left(x, x+\mu\hat\xi, u\right)
 \hat{\xi}\right]}{\mu}, \label{eq-diff-of-grad}
 \end{align}
where we used the fact that $\mathbb{E}_{\hat\xi} \left[g(x)\hat\xi\right] = 0$ as $\xi$ has mean zero and the identity 
$
\mathbb{E}_{\hat\xi} \left[ \left\langle u, \hat\xi \right\rangle \hat\xi \right] = u,
$
which follows from the fact $\mathbb{E}\left(\hat\xi {\hat\xi}^\top\right)= I_d$. Using \eqref{subgrad-bound-3} and \eqref{eq-diff-of-grad}, we obtain for $\mu>0$,
\begin{equation} 
\left\|\nabla g_0(x) - u\right\| \leq\mathbb{E}_{\hat{\xi}} \left[ \frac{\left| o\left(x, x+\mu\hat\xi,u\right)\right| 
}{\mu^{2}\|\hat\xi\|^{2}} \mu\| \hat{\xi}\|^{3} \right]\\
 \leq  C_{g} \mu \mathbb{E}_{\hat\xi}\left(\|\hat\xi\|^{3}\right).
\end{equation}
for any $u\in\partial g(x)$. 
Finally, we can write $\hat{\xi}=(\hat{\xi}_{1},\ldots,\hat{\xi}_{d})$, where $\xi_{i}$ 
are i.i.d. $\mathcal{N}(0,1)$ random variables and by Jensen's inequality,
\begin{align}
\mathbb{E}\left[\Vert\hat{\xi}\Vert^{3}\right]
&=\mathbb{E}\left[\left(|\hat{\xi}_{1}|^{2}+\cdots+|\hat{\xi}_{d}|^{2}\right)^{3/2}\right]
\leq\left(\mathbb{E}\left[\left(|\hat{\xi}_{1}|^{2}+\cdots+|\hat{\xi}_{d}|^{2}\right)^{2}\right]\right)^{3/4}
\nonumber
\\&\leq\left(d\mathbb{E}\left[|\hat{\xi}_{1}|^{4}+\cdots+|\hat{\xi}_{d}|^{4}\right]\right)^{3/4}
=(3d^{2})^{3/4}=3^{3/4}d^{3/2}.
\end{align}
This completes the proof.
\hfill $\Box$


\subsection*{Proof of Proposition~\ref{prop-when-assump-hold}}
Since $b(x)= -\nabla U_0(x) e^{U(x)-U_0(x)}$ we can write $b(x) = -\nabla U_0(x) + e(x)$ where $e(x) = -U_0(x) (e^{g(x)-g_0(x)}-1)$. Since $g$ is weakly convex, 
$g(x)$ is differentiable in a generalized sense (in the sense of Norkin) \citep{norkin1980generalized, zhu2023distributionally}. Then, by \cite[Theorem A.1]{norkin1980generalized}, $e(x)$ is also 
generalized differentiable and by the chain rule on a path for generalized differentiable functions \cite[Theorem A.3]{gurbuzbalaban2022stochastic}, we can write 
\begin{equation} 
e(x) - e(y) = \int_{t=0}^1 \left\langle s(x+t(y-x)), y-x \right\rangle dt, \label{path-integral}
\end{equation}
where $s(x+t(y-x))$ denotes any element of the subdifferential of $e$ at $x+t(y-x)$. Furthermore, by the chain rule \cite[Theorem A.1]{gurbuzbalaban2022stochastic}, $e(x)$ admits the subdifferential
$$\partial e(x) = -\nabla U_0(x)\left(e^{g(x)-g_0(x)}-1\right) -U_0(x) \cdot e^{g(x)-g_0(x)}\cdot (\partial g(x)-\nabla g_0(x)).$$ 
By the assumptions and \eqref{path-integral}, $\|s(x+t(y-x))\| = o(\mu)$, $\|e(x)-e(y)\|=o(\mu)\|x-y\|$ and
$|\langle e(x) - e(y) , x-y \rangle| = \|x-y\|^2 o(\mu)$. 
Therefore
\begin{align} 
\langle b(x)-b(y), x- y \rangle &= \langle -\nabla U_0(x) + \nabla U_0(y) + e(x) - e(y), x- y \rangle\nonumber \\
&= \langle -\nabla U_0(x) + \nabla U_0(y) , x- y \rangle + \langle e(x) - e(y), x- y \rangle.  
\end{align}
Since $U_0$ is $c_0$-strongly convex and $L_0$-smooth, we then have
\begin{equation} 
\langle b(x)-b(y), x- y \rangle  \leq (- c_0 + o(\mu)) \|x-y\|^2, 
\end{equation}
which implies that \eqref{discrete:condition:1} holds for $\mu>0$ small enough.
Also, we have 
$\| b(x)-b(y)\| = \|\nabla U_0(x) - \nabla U_0(y)\| + \|e(x)-e(y)\| \leq (L_0 + o(\mu)) \|x-y\|$. Therefore, \eqref{discrete:condition:1} holds when $\mu$ is sufficiently small. Similarly, $\sigma(x):=e^{(U(x)-U_{0}(x))/2}=e^{(g(x)-g_{0}(x))/2}$ is generalized differentiable and 
\begin{equation} 
\partial \sigma(x) = \frac{1}{2}e^{(g(x)-g_{0}(x))/2} \cdot (\partial g(x) - \nabla g_0(x)),\label{eq-subdiff-sigma}
\end{equation}
and we can write
\begin{equation}\sigma(x)-\sigma(y) = \int_{t=0}^1\langle s_\sigma(x+t(y-x)), y-x\rangle dt,\label{eq-sigma-path-integral}
\end{equation}
where $s_\sigma(x+t(y-x))$ is any element of the subdifferential $\partial \sigma(x+t(y-x))$. Furthermore, by Lemma~\ref{lem:CP:bound}, Lemma~\ref{ineq-small-grad-error} and \eqref{eq-subdiff-sigma}, 
$\sup_{z\in\mathbb{R}^{d}} \{\|s_\sigma(z)\| : s_\sigma (z)\in \partial \sigma(x)\} = \mathcal{O}(\mu)$.
Therefore, by a similar argument, from \eqref{eq-sigma-path-integral}, we have 
\begin{equation}
	\Vert\sigma(x)I_{d}-\sigma(y)I_{d}\Vert_{\mathrm{HS}}\leq\mathcal{O}(\mu) \Vert x-y\Vert,\qquad\text{for any $x,y\in\mathbb{R}^{d}$},
\end{equation}
which implies \eqref{discrete:condition:3} holds when $\mu$ is small enough. We then conclude that Assumption~\ref{assump:b:sigma} holds and this completes the proof.
\hfill $\Box$

\subsection*{Proof of Lemma~\ref{lem:exposquared:CP}}

    We have $U(x)-U_0(x)\leq K\mu\sqrt{d}$ by Lemma~\ref{lem:CP:bound}, which implies 
    \begin{equation}\label{lem:exposquared:1:CP}
         e^{U(x)-U_0(x)} \leq e^{K\mu\sqrt{d}}.
    \end{equation}
    To derive Eq. \eqref{lem:exposquared:2:CP}, we can first compute that
    \begin{align*}
        \mathbb{E}\left[\left| e^{U(x)-\tilde{U}_{0}(x)}-e^{U(x)-U_{0}(x)}\right|^2\right] &= e^{2U(x)-2U_{0}(x)}\mathbb{E}\left[\left| e^{U_0(x)-\tilde{U}_{0}(x)}-1\right|^2\right] \\
        &= e^{2U(x)-2U_{0}(x)}\mathbb{E}\left[ e^{2U_0(x)-2\tilde{U}_{0}(x)}-2e^{U_0(x)-\tilde{U}_{0}(x)}+1\right].
    \end{align*}
    By Jensen's inequality, we get 
    \begin{equation}
        \mathbb{E}\left[-2e^{U_0(x)-\tilde{U}_{0}(x)}+1\right] \leq -2e^{\mathbb{E}[U_0(x)-\tilde{U}_0(x)]}+1 = -1, 
    \end{equation}
    which together with Eq. \eqref{lem:exposquared:1:CP} implies that:
    \begin{equation}\label{Eq:expo3:CP}
        \mathbb{E}\left[\left| e^{U(x)-\tilde{U}_{0}(x)}-e^{U(x)-U_{0}(x)}\right|^2\right] \leq e^{2K\mu\sqrt{d} }\mathbb{E}\left[ e^{2U_0(x)-2\tilde{U}_{0}(x)}-1\right].
    \end{equation}
    We next derive an upper bound for $\mathbb{E}\left[ e^{2U_0(x)-2\tilde{U}_{0}(x)}\right]$. Denote $\xi_i$'s as the random noise in approximating $\tilde{U}_0(x)$, where $\xi_i\sim\mathcal{N}(0,I_{d})$ are i.i.d. over $i$. For $\xi\sim\mathcal{N}(0,I_{d})$, using Lemma~\ref{lem:CP:bound}, we have
    \begin{align*}
        \left|U_0(x)-\tilde{U}_0(x)\right|& \leq \left|U_0(x)-U(x)\right|+\left|U(x)-\tilde{U}_0(x)\right|\\
        &=\left|U_0(x)-U(x)\right|+\frac{1}{N}\left|\sum_{i=1}^{N}(g(x)-g(x+\mu\xi_i))\right| \\
        & \leq K\mu\sqrt{d}+\frac{1}{N}\sum_{i=1}^{N}K\mu\mathbb{E}\Vert\xi_i\Vert\leq 2K\mu\sqrt{d}.
    \end{align*}
    One can check that for $|x| \leq 4K\mu\sqrt{d}$, $e^x \leq 1+e^{4K\mu\sqrt{d}}|x|$. Also, using Jensen's inequality, we get
    \begin{align*}
        \mathbb{E}\left[ e^{2U_0(x)-2\tilde{U}_{0}(x)}\right] &\leq 1+\mathbb{E}\left[e^{4K\mu\sqrt{d}}\left|2U_0(x)-2\tilde{U}_0(x)\right|\right] \\
        &\leq 1+\mathbb{E}\left[e^{4K\mu\sqrt{d}}\right]\mathbb{E}\left[\left(2U_0(x)-2\tilde{U}_0(x)\right)^2\right]^{\frac{1}{2}} \\
        &\leq 1+2\cdot e^{4K\mu\sqrt{d}}\cdot \mathbb{E}\left[\left(U_0(x)-\tilde{U}_0(x)\right)^2\right]^{\frac{1}{2}},
    \end{align*}
    which can be combined with the inequality in Eq.~\eqref{Eq:expo3:CP} to obtain:
    \begin{equation*}
     \mathbb{E}\left[\left| e^{U(x)-\tilde{U}_{0}(x)}-e^{U(x)-U_{0}(x)}\right|^2\right] \leq 2\cdot  e^{6K\mu \sqrt{d}}\cdot\mathbb{E}\left[(U_0(x)-\tilde{U}_0(x))^2\right]^{\frac{1}{2}}.
    \end{equation*}
    We can also derive that
    \begin{align*}      \mathbb{E}\left[\left(U_0(x)-\tilde{U}_0(x)\right)^2\right] &= \mathbb{E}\left[U_0(x)-\tilde{U}_0(x)\right]^2 + \mathrm{Var}\left(U_0(x)-\tilde{U}_0(x)\right) \\
        & = \mathrm{Var}\left(\tilde{U}_0(x)\right) = \mathrm{Var}\left(\frac{1}{N}\sum_{i=1}^{N}g(x+\mu\xi_i)\right)= \frac{1}{N}\mathrm{Var}(g(x+\mu\xi)).
    \end{align*}
    We can then decompose and bound the variance of $g(x+\mu\xi)$ as
    \begin{align*}
        \mathrm{Var}(g(x+\mu\xi)) &= \mathbb{E}\left[\left(g(x+\mu\xi)-\mathbb{E}[g(x+\mu\xi)]\right)^2\right] \\
        & = \mathbb{E}\left[\left(g(x+\mu\xi)-g(x)+g(x)+\mathbb{E}[g(x+\mu\xi)]\right)^2\right] \\
        & \leq 2\mathbb{E}\left[(g(x+\mu\xi)-g(x))^2\right]+2\mathbb{E}\left[(g(x)-\mathbb{E}[g(x+\mu\xi)])^2\right] \\
        & \leq 2K^2\mu^{2}\mathbb{E}[\|\xi\|^{2}]+2K^2\mu^{2}\mathbb{E}[\|\xi\|]^2 
        \leq 4K^2\mu^{2}\mathbb{E}[\|\xi\|^{2}]=4K^{2}\mu^{2}d.
    \end{align*}
    Combining the above inequalities, we get
    \begin{align*}
        \mathbb{E}\left[\left| e^{U(x)-\tilde{U}_{0}(x)}-e^{U(x)-U_{0}(x)}\right|^2\right] \leq \frac{2}{\sqrt{N}}\cdot  e^{6K\mu\sqrt{d}}\cdot\,\mathrm{Var}(g(x+\mu\xi))^{\frac{1}{2}} \leq \frac{4K\mu\sqrt{d}}{\sqrt{N}}\cdot  e^{6K\mu\sqrt{d}},
    \end{align*}
    which completes the proof.
\hfill $\Box$


\subsection*{Proof of Theorem~\ref{thm:MC:CP}}

Using the notation of $\xi_i$'s as the random noise in approximating $\tilde{U}_0(x)$ and $\hat{\xi}_i$'s as the noise for $\nabla \tilde{U}_0(x)$, with $\xi_i\sim\mathcal{N}(0,I_{d})$ and $\hat{\xi}_i \sim\mathcal{N}(0,I_{d})$ being i.i.d. over $i$, we first prove the inequality in Eq. \eqref{bound:b:CP}. We have
\begin{align}\label{Eq:b:MC:CP}
    \mathbb{E}\left[\left\Vert\tilde{b}(x)-b(x)\right\Vert^2\right] &= \mathbb{E}\left[\left\Vert\nabla \tilde{U}_{0}(x)e^{U(x)-\tilde{U}_{0}(x)}-\nabla U_{0}(x)e^{U(x)-U_{0}(x)}\right\Vert^2\right] \nonumber\\
    &\leq 2\mathbb{E}\left[\left\Vert\nabla \tilde{U}_{0}(x)e^{U(x)-\tilde{U}_{0}(x)}-\nabla \tilde{U}_{0}(x)e^{U(x)-U_{0}(x)}\right\Vert^2\right] \nonumber\\
    &\qquad +2\mathbb{E}\left[\left\Vert\nabla \tilde{U}_{0}(x)e^{U(x)-U_{0}(x)}-\nabla U_{0}(x)e^{U(x)-U_{0}(x)}\right\Vert^2\right] \nonumber\\
    &= 2\mathbb{E}\left[\left\Vert\nabla \tilde{U}_{0}(x)\right\Vert^2\right] \mathbb{E}\left[\left| e^{U(x)-\tilde{U}_{0}(x)}-e^{U(x)-U_{0}(x)}\right|^2\right] \nonumber\\
    &\qquad +2e^{2U(x)-2U_0(x)}\mathbb{E}\left[\left\Vert \nabla \tilde{U}_0(x)-\nabla U_0(x)\right\Vert^2\right].
\end{align}
We then have the following inequality:
\begin{align}
    \mathbb{E}\left[\left\Vert\nabla \tilde{U}_{0}(x)\right\Vert^2\right] & = \mathbb{E}\left[\left\Vert\nabla f(x)+\frac{1}{\mu N}\sum_{i=1}^{N}\hat{\xi}_ig\left(x+\mu\hat{\xi}_i\right)\right\Vert^2\right] \nonumber
    \\
    &\leq 2\Vert\nabla f(x)\Vert^{2}+ 2\mathbb{E}\left[\left\Vert\frac{1}{\mu N}\sum_{i=1}^{N}\hat{\xi}_ig\left(x+\mu\hat{\xi}_i\right)\right\Vert^2\right]\nonumber
    \\
    &\leq 2\Vert\nabla f(x)\Vert^{2}+\frac{2}{\mu^2 N^2}\mathbb{E}\left[N\sum_{i=1}^{N}\left\Vert \hat{\xi}_ig\left(x+\mu\hat{\xi}_i\right)\right\Vert^2\right]\nonumber
    \\&= 2\Vert\nabla f(x)\Vert^{2}+\frac{2}{\mu^2} \mathbb{E}\left[\Vert\hat{\xi} g(x+\mu\hat{\xi})\Vert^2\right],\label{grad:tilde:U:0}
\end{align}	
where we used Cauchy-Schwarz inequality and the last equality is due to the independence of $\xi_i$'s. Here, we need to use Lemma~\ref{lem:exposquared:CP} to get the bounds for other terms. Combining the above inequality and Lemma~\ref{lem:exposquared:CP} under the choice of $\mu\leq\frac{1}{6K\sqrt{d}}$, we get
\begin{align}
    &\mathbb{E}\left[\left\Vert\nabla \tilde{U}_{0}(x)\right\Vert^2\right] \mathbb{E}\left[\left| e^{U(x)-\tilde{U}_{0}(x)}-e^{U(x)-U_{0}(x)}\right|^2\right]
    \nonumber
    \\&\leq \frac{(1+e)}{\sqrt{N}}\left(2\Vert\nabla f(x)\Vert^{2}+\frac{2}{\mu^2}\mathbb{E}\left[\Vert\hat{\xi} g(x+\mu\hat{\xi})\Vert^2\right]\right).\label{Eq:b:firstterm:CP}
\end{align}
We can also compute that
\begin{align}\label{Eq:MC:grad:CP}
    &\mathbb{E}\left[\left\Vert \nabla \tilde{U}_0(x)-\nabla U_0(x)\right\Vert^2\right] \nonumber
    \\&= \mathbb{E}\left[\left\Vert \frac{1}{\mu N}\sum_{i=1}^{N}\left(\hat{\xi}_ig\left(x+\mu\hat{\xi}_i\right)- \mathbb{E}\left[\hat{\xi}_ig\left(x+\mu\hat{\xi}_i\right)\right]\right)\right\Vert^2\right] \nonumber\\
    & = \frac{1}{\mu^2N^2}\mathbb{E}\left[N\left\Vert \left(\hat{\xi}_1g\left(x+\mu\hat{\xi}_1\right)- \mathbb{E}\left[\hat{\xi}_1g\left(x+\mu\hat{\xi}_1\right)\right]\right)\right\Vert^2\right] \nonumber\\
    &\qquad+\frac{1}{\mu^2N^2}\mathbb{E}\Bigg[N(N-1)\Bigg\langle \hat{\xi}_1g\left(x+\mu\hat{\xi}_1\right)- \mathbb{E}\left[\hat{\xi}_1g\left(x+\mu\hat{\xi}_1\right)\right],
    \nonumber
    \\
&\qquad\qquad\qquad\qquad\qquad\qquad\qquad\qquad\hat{\xi}_2g\left(x+\mu\hat{\xi}_2\right)- \mathbb{E}\left[\hat{\xi}_2g\left(x+\mu\hat{\xi}_2\right)\right]\Bigg\rangle\Bigg],
\end{align}
where we can compute that the second
term on the right hand side of \eqref{Eq:MC:grad:CP} 
is zero due to the independence of $\hat{\xi}_i$'s. Hence, we can simplify the equality in Eq.~\eqref{Eq:MC:grad:CP} and get
\begin{align}\label{Eq:b:secondterm:CP}
    & e^{2U(x)-2U_0(x)}\mathbb{E}\left[\left\Vert \nabla \tilde{U}_0(x)-\nabla U_0(x)\right\Vert^2\right] \nonumber\\
    &\qquad = \frac{1}{\mu^2N}e^{2U(x)-2U_0(x)}\mathbb{E}\left[\left\Vert \hat{\xi} g\left(x+\mu\hat{\xi}\right)- \mathbb{E}\left[\hat{\xi} g\left(x+\mu\hat{\xi}\right)\right]\right\Vert^2\right] \nonumber\\
    &\qquad \leq \frac{e^{2K\mu\sqrt{d}}}{\mu^2\sqrt{N}}\mathbb{E}\left[\left\Vert \hat{\xi} g\left(x+\mu\hat{\xi}\right)- \mathbb{E}\left[\hat{\xi} g\left(x+\mu\hat{\xi}\right)\right]\right\Vert^2\right] \nonumber\\
    &\qquad \leq \frac{1}{\mu^2\sqrt{N}}\left(1+e\right)\mathbb{E}\left[\left\Vert \hat{\xi} g\left(x+\mu\hat{\xi}\right)- \mathbb{E}\left[\hat{\xi} g\left(x+\mu\hat{\xi}\right)\right]\right\Vert^2\right],
\end{align}
where the inequality above is due to Lemma~\ref{lem:CP:bound}, $K\mu \leq \sfrac{1}{6\sqrt{d}}$ and $N\geq\sqrt{N}$ for $N\geq 1$. Now we apply the result in Eq. \eqref{Eq:b:firstterm:CP} and Eq. \eqref{Eq:b:secondterm:CP} to the inequality in Eq. \eqref{Eq:b:MC:CP} to get 
\begin{equation}
\mathbb{E}\left[\left\Vert\tilde{b}(x)-b(x)\right\Vert^2\Big|x\right] \leq \frac{1}{\mu^{2}\sqrt{N}}\psi_b(x):= \psi_1(x)+\psi_2(x), 
\end{equation}
where
\begin{align*}
        \psi_1(x) &:=4(1+e)\left(\mu^{2}\left\Vert\nabla f(x)\right\Vert^{2}+\mathbb{E}\left[\left\Vert\hat{\xi} g\left(x+\mu\hat{\xi}\right)\right\Vert^2\right]\right),\\
        \psi_2(x) &:=2\left(1+e\right)\mathbb{E}\left[\left\Vert \hat{\xi} g\left(x+\mu\hat{\xi}\right)- \mathbb{E}\left[\hat{\xi} g\left(x+\mu\hat{\xi}\right)\right]\right\Vert^2\right].
\end{align*}
For the bound in Eq. \eqref{bound:sigma:CP}, by applying the same set-up as Lemma~\ref{lem:exposquared:CP}, one can derive a similar result:
\begin{align*}
    \mathbb{E}\left[\left| e^{(U(x)-\tilde{U}_{0}(x))/2}-e^{(U(x)-U_{0}(x))/2}\right|^2\right] \leq \frac{2K\mu\sqrt{d}}{\sqrt{N}}\cdot  e^{3K\mu\sqrt{d}}\leq \frac{1}{3\sqrt{N}}\left(1+\frac{1}{2}e\right).
\end{align*}
Hence, we have
    \begin{align}
       \mathbb{E}\left[\left|\tilde{\sigma}(x)-\sigma(x)\right|^2\Big|x\right] \leq \frac{1}{\sqrt{N}}B, 
    \end{align}
    where
$B:=\frac{1}{3}\left(1+\frac{1}{2}e\right)$.
Finally, we can compute that
\begin{align}
\psi_{1}(x)
&\leq 4(1+e)\left(\mu^{2}\Vert\nabla f(x)\Vert^{2}+\mathbb{E}\left[\left\Vert\hat{\xi} g\left(x+\mu\hat{\xi}\right)\right\Vert^2\right]\right)
\nonumber
\\
&\leq 4(1+e)\left(\mu^{2}\left\Vert\nabla f(x)-\nabla f(x_{\ast}^{f})\right\Vert^{2}+\mathbb{E}\left[\left\Vert\hat{\xi} g\left(x+\mu\hat{\xi}\right)\right\Vert^2\right]\right)\nonumber
\\
&\leq 4(1+e)\left(\mu^{2}L_{f}^{2}\left\Vert x-x_{\ast}^{f}\right\Vert^{2}
+2\mathbb{E}\left[\left\Vert\hat{\xi} g\left(x+\mu\hat{\xi}\right)-\hat{\xi}g(x)\right\Vert^2\right]
+2\mathbb{E}\left[\Vert\hat{\xi}g(x)\Vert^2\right]\right)\nonumber
\\
&\leq 4(1+e)\left(2\mu^{2}L_{f}^{2}\Vert x\Vert^{2}
+2\mu^{2}L_{f}^{2}\Vert x_{\ast}^{f}\Vert^{2}
+2K^{2}\mu^{2}\mathbb{E}\left[\Vert\hat{\xi}\Vert^4\right]
+2(g(x))^{2}\mathbb{E}\left[\Vert\hat{\xi}\Vert^2\right]\right)\nonumber
\\
&\leq 4(1+e)\left(2\mu^{2}L_{f}^{2}\Vert x\Vert^{2}
+2\mu^{2}L_{f}^{2}\Vert x_{\ast}^{f}\Vert^{2}
+2K^{2}\mu^{2}(3d^{2})
+2(g(x))^{2}d\right)\nonumber
\\
&\leq 4(1+e)\left(2\mu^{2}L_{f}^{2}\Vert x\Vert^{2}
+2\mu^{2}L_{f}^{2}\Vert x_{\ast}^{f}\Vert^{2}
+2K^{2}\mu^{2}(3d^{2})
+4(g(0))^{2}d+4K^{2}d\Vert x\Vert^{2}\right),\label{psi:1:upper:bound}
\end{align}
where $x_{\ast}^{f}$ is the unique minimizer of $f$ so that $\nabla f(x_{\ast}^{f})=0$.

Similarly, we can compute that
\begin{align*}
\psi_{2}(x)
&\leq
4\left(1+e\right)\mathbb{E}\left[\left(\left\Vert \hat{\xi} g\left(x+\mu\hat{\xi}\right)\right\Vert^{2}+\left\Vert \mathbb{E}\left[\hat{\xi} g\left(x+\mu\hat{\xi}\right)\right]\right\Vert^2\right)\right]
\\
&\leq
4\left(1+e\right)\Bigg(2\mathbb{E}\left\Vert \hat{\xi} g\left(x+\mu\hat{\xi}\right)-\hat{\xi} g(x)\right\Vert^{2}
+2\mathbb{E}\left\Vert\hat{\xi}g(x)\right\Vert^{2}
\\
&\qquad\qquad\qquad\qquad\qquad\qquad
+\left\Vert \mathbb{E}\left[\hat{\xi} g\left(x+\mu\hat{\xi}\right)\right]-\mathbb{E}\left[\hat{\xi}g(x)\right]\right\Vert^2\Bigg)
\\
&\leq
4\left(1+e\right)\left(2\mu^{2}K^{2}\mathbb{E}\left[\left\Vert \hat{\xi} \right\Vert^{4}\right]
+2(g(x))^{2}\mathbb{E}\left\Vert\hat{\xi}\right\Vert^{2}
+\mu^{2}K^{2}\left(\mathbb{E}\Vert\hat{\xi}\Vert^{2}\right)^{2}\right)
\\
&\leq
4\left(1+e\right)\left(2\mu^{2}K^{2}(3d^{2})+2(g(x))^{2}d+\mu^{2}K^{2}d^{2}\right)
\\
&\leq
4\left(1+e\right)\left(2\mu^{2}K^{2}(3d^{2})+4(g(0))^{2}d+4K^{2}d\Vert x\Vert^{2}+\mu^{2}K^{2}d^{2}\right),
\end{align*}
where we used $\mathbb{E}[\hat{\xi}]=0$.
This completes the proof.
\hfill $\Box$


\subsection*{Proof of Lemma~\ref{lem:L2}}
We can compute that
\begin{align*}
\tilde{x}_{k+1}-x_{\ast} =\tilde{x}_{k}-x_{\ast}
+\eta\tilde{b}(\tilde{x}_{k})+\sqrt{2\eta}\tilde{\sigma}(\tilde{x}_{k})\xi_{k+1},
\end{align*}
where $x_{\ast}$ is the minimizer of $U_{0}$ so that $b(x_{\ast})=0$.
Therefore,
\begin{align*}
\mathbb{E}\Vert\tilde{x}_{k+1}-x_{\ast}\Vert^{2} =\mathbb{E}\left\Vert\tilde{x}_{k}-x_{\ast}
+\eta\tilde{b}(\tilde{x}_{k})\right\Vert^{2}+2\eta\mathbb{E}(\tilde{\sigma}(\tilde{x}_{k}))^{2}\mathbb{E}\Vert\xi_{k+1}\Vert^{2},
\end{align*}
and moreover
\begin{align*}
\mathbb{E}\left\Vert\tilde{x}_{k}-x_{\ast}
+\eta\tilde{b}(\tilde{x}_{k})\right\Vert^{2}
=
\mathbb{E}\Vert\tilde{x}_{k}-x_{\ast}\Vert^{2}
+\eta^{2}\mathbb{E}\Vert\tilde{b}(\tilde{x}_{k})\Vert^{2}
+2\eta\mathbb{E}\langle\tilde{x}_{k}-x_{\ast},\tilde{b}(\tilde{x}_{k})\rangle.
\end{align*}
Next, we recall that
$\tilde{b}(x):=-\nabla \tilde{U}_{0}(x)e^{U(x)-\tilde{U}_{0}(x)}$ and $\tilde{\sigma}(x):=e^{(U(x)-\tilde{U}_{0}(x))/2}$.
Therefore, 
\begin{equation}
\mathbb{E}\left[(\tilde{\sigma}(\tilde{x}_{k}))^{2}\right]
=
\mathbb{E}\left[e^{U(\tilde{x}_{k})-\tilde{U}_{0}(\tilde{x}_{k})}\right]\leq
e^{3K\mu\sqrt{d}},
\end{equation}
where we used $U(x)-U_{0}(x)\leq K\mu\sqrt{d}$
and $U_{0}(x)-\tilde{U}_{0}(x)\leq 2K\mu\sqrt{d}$
from the proof of Lemma~\ref{lem:exposquared:CP},
and
\begin{align*}
\mathbb{E}\left\Vert\tilde{b}(\tilde{x}_{k})\right\Vert^{2}
&\leq
e^{6K\mu\sqrt{d}}\mathbb{E}\left\Vert\nabla\tilde{U}_{0}(\tilde{x}_{k})\right\Vert^{2}
\\
&\leq
\frac{2e^{6K\mu\sqrt{d}}}{\mu^{2}}\left(\mu^{2}\mathbb{E}\Vert\nabla f(\tilde{x}_{k})\Vert^{2}+ \mathbb{E}\left[\left\Vert\hat{\xi} g\left(\tilde{x}_{k}+\mu\hat{\xi}\right)\right\Vert^2\right]\right)
\\
&\leq
\frac{2e^{6K\mu\sqrt{d}}}{\mu^{2}}
\left(\left(2\mu^{2}L_{f}^{2}+4K^{2}d\right)\mathbb{E}\Vert\tilde{x}_{k}\Vert^{2}
+2\mu^{2}L_{f}^{2}\Vert x_{\ast}^{f}\Vert^{2}
+2K^{2}\mu^{2}(3d^{2})
+4(g(0))^{2}d\right)
\\
&\leq
\frac{2e^{6K\mu\sqrt{d}}}{\mu^{2}}
\Big(\left(4\mu^{2}L_{f}^{2}+8K^{2}d\right)\mathbb{E}\Vert\tilde{x}_{k}-x_{\ast}\Vert^{2}
+\left(4\mu^{2}L_{f}^{2}+8K^{2}d\right)\Vert x_{\ast}\Vert^{2}
\\
&\qquad\qquad\qquad\qquad\qquad\qquad
+2\mu^{2}L_{f}^{2}\Vert x_{\ast}^{f}\Vert^{2}
+2K^{2}\mu^{2}(3d^{2})
+4(g(0))^{2}d\Big),
\end{align*}
where we applied \eqref{grad:tilde:U:0} and \eqref{psi:1:upper:bound} in the proof of Theorem~\ref{thm:MC:CP}.

Next, by Assumption~\ref{assump:b:sigma} and Theorem~\ref{thm:MC:CP}, we can compute that
\begin{align*}
&\mathbb{E}\left\langle\tilde{x}_{k}-x_{\ast},\tilde{b}(\tilde{x}_{k})\right\rangle
\\
&=
\mathbb{E}\left\langle\tilde{x}_{k}-x_{\ast},b(\tilde{x}_{k})\right\rangle
+\mathbb{E}\left\langle\tilde{x}_{k}-x_{\ast},\tilde{b}(\tilde{x}_{k})-b(\tilde{x}_{k})\right\rangle
\\
&=
\mathbb{E}\left\langle\tilde{x}_{k}-x_{\ast},b(\tilde{x}_{k})-b(x_{\ast})\right\rangle
+\mathbb{E}\left\langle\tilde{x}_{k}-x_{\ast},\tilde{b}(\tilde{x}_{k})-b(\tilde{x}_{k})\right\rangle
\\
&\leq\mathbb{E}\left\langle\tilde{x}_{k}-x_{\ast},b(\tilde{x}_{k})-b(x_{\ast})\right\rangle
+\left(\mathbb{E}\Vert\tilde{x}_{k}-x_{\ast}\Vert^{2}\right)^{1/2}
\left(\mathbb{E}\Vert\tilde{b}(\tilde{x}_{k})-b(\tilde{x}_{k})\Vert^{2}\right)^{1/2}
\\
&\leq
-m\mathbb{E}\Vert\tilde{x}_{k}-x_{\ast}\Vert^{2}
+\left(\mathbb{E}\Vert\tilde{x}_{k}-x_{\ast}\Vert^{2}\right)^{1/2}
\frac{1}{\mu N^{1/4}}(A_{1}\mathbb{E}\Vert\tilde{x}_{k}\Vert^{2}+A_{2})^{1/2}
\\
&\leq
-m\mathbb{E}\Vert\tilde{x}_{k}-x_{\ast}\Vert^{2}
+\left(\mathbb{E}\Vert\tilde{x}_{k}-x_{\ast}\Vert^{2}\right)^{1/2}
\frac{1}{\mu N^{1/4}}(2C_{1}\mathbb{E}\Vert\tilde{x}_{k}-x_{\ast}\Vert^{2}+2A_{1}\Vert x_{\ast}\Vert^{2}+A_{2})^{1/2}.
\end{align*}
Furthermore, we can compute that
\begin{align*}
&\left(\mathbb{E}\Vert\tilde{x}_{k}-x_{\ast}\Vert^{2}\right)^{1/2}
\frac{1}{\mu N^{1/4}}(2C_{1}\mathbb{E}\Vert\tilde{x}_{k}-x_{\ast}\Vert^{2}+2A_{1}\Vert x_{\ast}\Vert^{2}+A_{2})^{1/2}
\\
&=\left(\mathbb{E}\Vert\tilde{x}_{k}-x_{\ast}\Vert^{2}\right)^{1/2}
\frac{\sqrt{2A_{1}}}{\mu N^{1/4}}\left(\mathbb{E}\Vert\tilde{x}_{k}-x_{\ast}\Vert^{2}+\Vert x_{\ast}\Vert^{2}+\frac{A_{2}}{2A_{1}}\right)^{1/2}
\\
&\leq
\frac{\sqrt{2A_{1}}}{\mu N^{1/4}}\left(\mathbb{E}\Vert\tilde{x}_{k}-x_{\ast}\Vert^{2}+\Vert x_{\ast}\Vert^{2}+\frac{A_{2}}{2A_{1}}\right).
\end{align*}
Putting everything together, we have
\begin{align*}
&\mathbb{E}\Vert\tilde{x}_{k+1}-x_{\ast}\Vert^{2} 
\\
&\leq
\mathbb{E}\Vert\tilde{x}_{k}-x_{\ast}\Vert^{2}
+2\eta\mathbb{E}(\tilde{\sigma}(\tilde{x}_{k}))^{2}\mathbb{E}\Vert\xi_{k+1}\Vert^{2}
+\eta^{2}\mathbb{E}\Vert\tilde{b}(\tilde{x}_{k})\Vert^{2}
+2\eta\mathbb{E}\langle\tilde{x}_{k}-x_{\ast},\tilde{b}(\tilde{x}_{k})\rangle
\\
&\leq
\mathbb{E}\Vert\tilde{x}_{k}-x_{\ast}\Vert^{2}
+2\eta e^{3K\mu\sqrt{d}}d
+\eta^{2}\frac{2e^{6K\mu\sqrt{d}}}{\mu^{2}}
(4\mu^{2}L_{f}^{2}+8K^{2}d)\mathbb{E}\Vert\tilde{x}_{k}-x_{\ast}\Vert^{2}
\\
&\qquad
+\eta^{2}\frac{2e^{6K\mu\sqrt{d}}}{\mu^{2}}
\left((4\mu^{2}L_{f}^{2}+8K^{2}d)\Vert x_{\ast}\Vert^{2}
+2\mu^{2}L_{f}^{2}\Vert x_{\ast}^{f}\Vert^{2}
+2K^{2}\mu^{2}(3d^{2})
+4(g(0))^{2}d\right)
\\
&\qquad
-2\eta m\mathbb{E}\Vert\tilde{x}_{k}-x_{\ast}\Vert^{2}
+2\eta\frac{\sqrt{2A_{1}}}{\mu N^{1/4}}\mathbb{E}\Vert\tilde{x}_{k}-x_{\ast}\Vert^{2}
+2\eta\frac{\sqrt{2A_{1}}}{\mu N^{1/4}}\left(\Vert x_{\ast}\Vert^{2}+\frac{A_{2}}{2A_{1}}\right).
\end{align*}
Under the assumptions
$\eta\leq\frac{m\mu^{2}}{4e^{6K\mu\sqrt{\mu}d}\left(4\mu^{2}L_{f}^{2}+8K^{2}d\right)}$and
$N\geq\left(\frac{4\sqrt{2A_{1}}}{m\mu}\right)^{4}$,
it follows that
\begin{align*}
&\mathbb{E}\Vert\tilde{x}_{k+1}-x_{\ast}\Vert^{2} 
\leq
(1-\eta m)\mathbb{E}\Vert\tilde{x}_{k}-x_{\ast}\Vert^{2}
+2\eta e^{3K\mu\sqrt{d}}d
\\
&\qquad
+\eta^{2}\frac{2e^{6K\mu\sqrt{d}}}{\mu^{2}}
\left(\left(4\mu^{2}L_{f}^{2}+8K^{2}d\right)\Vert x_{\ast}\Vert^{2}
+2\mu^{2}L_{f}^{2}\Vert x_{\ast}^{f}\Vert^{2}
+2K^{2}\mu^{2}(3d^{2})
+4(g(0))^{2}d\right)
\\
&\qquad\qquad
+2\eta\frac{\sqrt{2A_{1}}}{\mu N^{1/4}}\left(\Vert x_{\ast}\Vert^{2}+\frac{A_{2}}{2A_{1}}\right),
\end{align*}
which implies that
\begin{align*}
&\mathbb{E}\Vert\tilde{x}_{k}-x_{\ast}\Vert^{2} 
\leq
(1-\eta m)^{k}\mathbb{E}\Vert\tilde{x}_{0}-x_{\ast}\Vert^{2}
+\frac{2}{m}e^{3K\mu\sqrt{d}}d
\\
&\qquad
+\frac{\eta}{m}\frac{2e^{6K\mu\sqrt{d}}}{\mu^{2}}
\left(\left(4\mu^{2}L_{f}^{2}+8K^{2}d\right)\Vert x_{\ast}\Vert^{2}
+2\mu^{2}L_{f}^{2}\Vert x_{\ast}^{f}\Vert^{2}
+2K^{2}\mu^{2}(3d^{2})
+4(g(0))^{2}d\right)
\\
&\qquad\qquad
+\frac{2}{m}\frac{\sqrt{2A_{1}}}{\mu N^{1/4}}\left(\Vert x_{\ast}\Vert^{2}+\frac{A_{2}}{2A_{1}}\right).
\end{align*}
This completes the proof.
\hfill $\Box$


\subsection*{Proof of Corollary~\ref{cor:tilde:b:b}}
This is an immediate consequence of Theorem~\ref{thm:MC:CP} and Lemma~\ref{lem:L2}.
\hfill $\Box$


\subsection*{Proof of Proposition~\ref{prop:MC:CP}}
    From the dynamics represented in Eq. \eqref{discrete:overdamped:k} and Eq. \eqref{discrete:MC:k}, at every iteration $k \in \mathbb{N}^*$, we can derive the following equality:
    \begin{align*}
        \mathbb{E}\left[\Vert\tilde{x}_{k+1}-x_{k+1}\Vert^2\right] &= \mathbb{E}\left[\left\Vert\tilde{x}_{k}-x_{k}+\eta(\tilde{b}(\tilde{x}_{k})-b(x_{k}))\right\Vert^2\right] +2\eta\mathbb{E}\left[\left\Vert\xi_{k+1}(\tilde{\sigma}(\tilde{x}_{k})-\sigma(x_{k}))\right\Vert^2\right]\\
        &\qquad+ \mathbb{E}\left[2\left\langle \tilde{x}_{k}-x_{k}+\eta(\tilde{b}(\tilde{x}_{k})-b(x_{k})),\xi_{k+1}(\tilde{\sigma}(\tilde{x}_{k})-\sigma(x_{k}))\right\rangle\right] \\
        &= \mathbb{E}\left[\Vert\tilde{x}_{k}-x_{k}\Vert^2\right]+\eta^2\mathbb{E}\left[\left\Vert\tilde{b}(\tilde{x}_{k})-b(x_{k})\right\Vert^2\right]\\
        &\qquad+2\eta\mathbb{E}\left[\left\langle \tilde{x}_{k}-x_{k},\tilde{b}(\tilde{x}_{k})-b(x_{k})\right\rangle\right] +2\eta d\mathbb{E}\left[\left\Vert\tilde{\sigma}(\tilde{x}_{k})-\sigma(x_{k})\right\Vert^2\right],
    \end{align*}
    where the second equality is due to $\xi_{k+1} \sim\mathcal{N}(0,I_d)$ being independent of $\tilde{b}(\tilde{x}_k)$, $b(x_k)$, $\tilde{\sigma}(\tilde{x}_k)$ and $\sigma(x_k)$. We can bound and further decompose the above expression as follows:
    \begin{align*}
        &\mathbb{E}\left[\Vert\tilde{x}_{k+1}-x_{k+1}\Vert^2\right]
        \\
        &\leq \mathbb{E}\left[\Vert\tilde{x}_{k}-x_{k}\Vert^2\right]+\eta^2\mathbb{E}\left[\left\Vert\tilde{b}(\tilde{x}_{k})-b(x_{k})\right\Vert^2\right]\\
        &\qquad+\eta\mathbb{E}\left[\Vert\tilde{x}_{k}-x_{k}\Vert^2\right]+\eta\mathbb{E}\left[\left\Vert\tilde{b}(\tilde{x}_{k})-b(x_{k})\right\Vert^2\right] +2\eta d\mathbb{E}\left[\left\Vert\tilde{\sigma}(\tilde{x}_{k})-\sigma(x_{k})\right\Vert^2\right] \\
        & = (1+\eta)\mathbb{E}\left[\Vert\tilde{x}_{k}-x_{k}\Vert^2\right]+(\eta^2+\eta)\mathbb{E}\left[\left\Vert\tilde{b}(\tilde{x}_{k})-b(\tilde{x}_k)+b(\tilde{x}_k)-b(x_{k})\right\Vert^2\right] \\
        &\qquad+2\eta d\mathbb{E}\left[|\tilde{\sigma}(\tilde{x}_{k})-\sigma(\tilde{x}_{k})+\sigma(\tilde{x}_{k})-\sigma(x_{k})|^2\right] \\
        &\leq (1+\eta)\mathbb{E}\left[\Vert\tilde{x}_{k}-x_{k}\Vert^2\right]+(2\eta^2+2\eta)\mathbb{E}\left[\left\Vert\tilde{b}(\tilde{x}_{k})-b(\tilde{x}_{k})\right\Vert^2\right] \\
        &+(2\eta^2+2\eta)\mathbb{E}\left[\Vert b(\tilde{x}_{k})-b(x_{k})\Vert^2\right]+4\eta d\mathbb{E}\left[\Vert\tilde{\sigma}(\tilde{x}_{k})-\sigma(\tilde{x}_{k})\Vert^2\right] +4\eta d\mathbb{E}\left[\Vert \sigma(\tilde{x}_{k})-\sigma(x_{k})\Vert^2\right].
    \end{align*}
    We can further bound the above terms as
    \begin{align*}
        \mathbb{E}\left[\Vert\tilde{x}_{k+1}-x_{k+1}\Vert^2\right] &\leq (1+\eta)\mathbb{E}\left[\Vert\tilde{x}_{k}-x_{k}\Vert^2\right]+(2\eta^2+2\eta)\frac{1}{\mu^2\sqrt{N}}A \\
        &\qquad+(2\eta^2+2\eta)L^2\mathbb{E}\left[\Vert\tilde{x}_{k}-x_{k}\Vert^2\right] +4\eta d\frac{1}{\sqrt{N}}B +4\eta d\alpha\mathbb{E}\left[\Vert\tilde{x}_{k}-x_{k}\Vert^2\right] \\
        &= (1+\eta+2\eta^2L^2+2\eta L^2+4\eta d\alpha)\mathbb{E}\left[\Vert\tilde{x}_{k}-x_{k}\Vert^2\right] \\
        &\qquad+(2\eta^2+2\eta)\frac{1}{\mu^2\sqrt{N}}A+4\eta d\frac{1}{\sqrt{N}}B.
    \end{align*}
    Using this inequality repeatedly for $k+1, k,\ldots, 1$, we arrive at
    \begin{align*}
        &\mathbb{E}\left[\Vert\tilde{x}_{k+1}-x_{k+1}\Vert^2\right] \leq \left(1+\eta+2\eta^2L^2+2\eta L^2+4\eta d\alpha\right)^{k+1}\mathbb{E}\left[\Vert\tilde{x}_{0}-x_{0}\Vert^2\right] \\
        &\qquad+\frac{(1+\eta+2\eta^2L^2+2\eta L^2+4\eta d\alpha)^{k+1}-1}{\eta+2\eta^2L^2+2\eta L^2+4\eta d\alpha}\left((2\eta^2+2\eta)\frac{1}{\mu^2\sqrt{N}}A+4\eta d\frac{1}{\sqrt{N}}B\right).
    \end{align*}
    Finally, we apply the property $\mathcal{W}_2(\nu_k,\tilde{\nu}_k) \leq \mathbb{E}\left[\Vert\tilde{x}_{k}-x_{k}\Vert^2\right]$ and $\tilde{x}_0=x_0$ to the above inequality to get
    \begin{equation*}
        \mathcal{W}_2(\nu_k,\tilde{\nu}_k) \leq \frac{(2\eta+2)\frac{1}{\mu^2\sqrt{N}}A+4d\frac{1}{\sqrt{N}}B}{1+2\eta L^2+2L^2+4d\alpha}\left((1+\eta+2\eta^2L^2+2\eta L^2+4\eta d\alpha)^{k}-1\right).
    \end{equation*}
    This completes the proof.
\hfill $\Box$


\subsection*{Proof of Theorem~\ref{thm:MC:final:CP}}
The results follows from Theorem~\ref{thm:discrete} and Proposition~\ref{prop:MC:CP}.
\hfill $\Box$


\subsection*{Proof of Corollary~\ref{cor:final:complexity:CP}}
Let us choose the parameters such that
$\sqrt{2}e^{-\beta k\eta}\mathcal{W}_{2}(\nu_{0},\pi) \leq \frac{\epsilon}{2}$,    $\sqrt{2}C\eta^{p_{2}-\frac{1}{2}} \leq \frac{\epsilon}{4}$ and $\tau\left(e^{\eta k\varrho}-1\right) \leq \frac{\epsilon}{4}$,
where we recall that $\beta=m-\alpha$ and $p_{2}=1$,
then it follows from Theorem~\ref{thm:MC:final:CP} 
that $\mathcal{W}_2(\tilde{\nu}_k,\pi) \leq \epsilon$, which yields the desired result.
\hfill $\Box$

\section{Supporting Lemmas}\label{sec:support:lemmas}

\begin{lemma}\label{lem:L2:contract}
	The anchored Langevin SDE \eqref{overdamped:SDE:modified}
	is contractive with rate $\beta:=m-\alpha>0$, i.e.
	\begin{equation}
		\mathbb{E}\Vert X_{t}-\tilde{X}_{t}\Vert^{2}
		\leq e^{-2\beta t}\mathbb{E}\Vert X_{0}-\tilde{X}_{0}\Vert^{2},
	\end{equation}
	where $X_{t},\tilde{X}_{t}$ are two solutions of \eqref{overdamped:SDE:modified}
	with synchronous coupling. 
\end{lemma}

\subsection*{Proof of Lemma~\ref{lem:L2:contract}}
	Since $X_{t},\tilde{X}_{t}$ are two solutions of \eqref{overdamped:SDE:modified}
	with synchronous coupling, we have
	\begin{align}
		dX_{t}=b(X_{t})dt+\sqrt{2}\sigma(X_{t})dW_{t},
		\qquad
		d\tilde{X}_{t}=b(\tilde{X}_{t})dt+\sqrt{2}\sigma(\tilde{X}_{t})dW_{t},
	\end{align}
	such that
	\begin{align*}
		\frac{d}{dt}\mathbb{E}\left\Vert X_{t}-\tilde{X}_{t}\right\Vert^{2}
		&=2\mathbb{E}\left[\left\langle b(X_{t})-b(\tilde{X}_{t}),X_{t}-\tilde{X}_{t}\right\rangle\right]
		+2\mathbb{E}\left\Vert\sigma(X_{t})I_{d}-\sigma(\tilde{X}_{t})I_{d}\right\Vert_{\mathrm{HS}}^{2}
		\\
		&\leq-2m\mathbb{E}\left\Vert X_{t}-\tilde{X}_{t}\right\Vert^{2}+2\alpha\mathbb{E}\left\Vert X_{t}-\tilde{X}_{t}\right\Vert^{2},
	\end{align*}
	which implies that
	$\mathbb{E}\left\Vert X_{t}-\tilde{X}_{t}\right\Vert^{2}
		\leq e^{-2(m-\alpha)t}\mathbb{E}\left\Vert X_{0}-\tilde{X}_{0}\right\Vert^{2}$.
	This completes the proof.
\hfill $\Box$

By adapting Lemma 4.3. in \cite{TaoMirror2021} to our setting, we immediately 
obtain the following lemma.

\begin{lemma}
	For any $t\geq 0$,
	$\mathbb{E}\left\Vert (X_{t}-X_{0})-\left(\tilde{X}_{t}-\tilde{X}_{0}\right)\right\Vert^{2}
		\leq C_{0}\mathbb{E}\Vert X_{0}-\tilde{X}_{0}\Vert^{2}t$,
	where $C_{0}:=4L$ and $X_{t},\tilde{X}_{t}$ are two solutions of \eqref{overdamped:SDE:modified}
	with synchronous coupling. 
\end{lemma}

Let $x_{\ast}$ be the minimizer of $U_{0}(x)$ such that $b(x_{\ast})=0$.
By adapting Lemma 4.4. in \cite{TaoMirror2021} to our setting, we immediately 
obtain the following lemma.

\begin{lemma}\label{lemma:cont:bound}
	For the anchored Langevin SDE \eqref{overdamped:SDE:modified}, for any $0<t\leq t_{0}:=\frac{1}{L^{2}+4\alpha}$, we have
	$\mathbb{E}\Vert X_{t}-X_{0}\Vert^{2}\leq\gamma t$,
	where
	$\gamma:=8(1+4\alpha)\mathbb{E}\Vert X_{0}\Vert^{2}+8(1+4\alpha)\Vert x_{\ast}\Vert^{2}+16\Vert\sigma(x_{\ast})I_{d}\Vert_{\mathrm{HS}}^{2}$.
\end{lemma}

By adapting Lemma 4.5. in \cite{TaoMirror2021} to our setting, we immediately 
obtain the following lemma.

\begin{lemma}
	The Euler discretization \eqref{discrete:dynamics} has local weak error at least
	of order $p_{1}:=3/2$ with maximum stepsize $\eta_{1}:=\frac{1}{L^{2}+4\alpha}$ and constants
	$C_{1}:=3L\sqrt{1+4\alpha}\left(\Vert x_{\ast}\Vert+\Vert\sigma(x_{\ast})I_{d}\Vert_{\mathrm{HS}}\right)$
    and $D_{1}:=2L\sqrt{1+4\alpha}$.
\end{lemma}

By adapting Lemma 4.6. in \cite{TaoMirror2021} to our setting, we immediately 
obtain the following lemma.

\begin{lemma}
	The Euler discretization \eqref{discrete:dynamics} has local strong error at least
	of order $p_{2}:=1$ with maximum stepsize $\eta_{2}:=\frac{1}{L^{2}+4\alpha}$ and constants
	$C_{2}:=7(1+4\alpha)\left(\Vert x_{\ast}\Vert+\Vert\sigma(x_{\ast})I_{d}\Vert_{\mathrm{HS}}\right)$ and
	$D_{2}:=5(1+4\alpha)$.
\end{lemma}

\vskip 0.2in
\bibliography{langevin}
	
\end{document}